\documentclass{article}

\PassOptionsToPackage{colorlinks=true, citecolor=blue, linkcolor=blue, urlcolor=black}{hyperref}

\usepackage{arxiv}
\usepackage[utf8]{inputenc}
\usepackage[T1]{fontenc}
\usepackage{url}
\usepackage{booktabs}
\usepackage{nicefrac}
\usepackage{microtype}
\usepackage{lipsum}
\usepackage{graphicx}
\usepackage{natbib}
\usepackage{doi}
\usepackage{algorithm}
\usepackage{algorithmicx}
\usepackage{multirow}
\usepackage{multicol}
\usepackage{amsmath,amssymb,amsfonts}
\usepackage{amsthm}
\usepackage{mathrsfs}
\usepackage{xcolor}
\usepackage{float}

\title{Multi-frequency wavefield solutions for variable velocity models using meta-learning enhanced low-rank physics-informed neural network}

\author{ \href{https://orcid.org/0000-0001-8868-7967}{\includegraphics[scale=0.06]{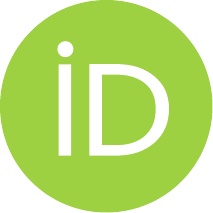}\hspace{1mm}Shijun~Cheng}\\
	Division of Physical Science and Engineering\\
	King Abdullah University of Science and Technology\\
	Thuwal 23955-6900, Saudi Arabia \\
	\texttt{sjcheng.academic@gmail.com} \\
        \And
	\href{https://orcid.org/0000-0002-9363-9799}{\includegraphics[scale=0.06]{orcid.pdf}\hspace{1mm}Tariq~Alkhalifah} \\
	Division of Physical Science and Engineering\\
	King Abdullah University of Science and Technology\\
	Thuwal 23955-6900, Saudi Arabia \\
	\texttt{tariq.alkhalifah@kaust.edu.sa} \\
}

\renewcommand{\shorttitle}{Meta-LRPINN}

\hypersetup{
pdftitle={A template for the arxiv style},
pdfsubject={q-bio.NC, q-bio.QM},
pdfauthor={David S.~Hippocampus, Elias D.~Striatum},
pdfkeywords={First keyword, Second keyword, More},
}

\begin{document}
\maketitle

\begin{abstract}
Physics-informed neural networks (PINNs) face significant challenges in modeling multi-frequency wavefields in complex velocity models due to their slow convergence, difficulty in representing high-frequency details, and lack of generalization to varying frequencies and velocity scenarios. To address these issues, we propose Meta-LRPINN, a novel framework that combines low-rank parameterization using singular value decomposition (SVD) with meta-learning and frequency embedding. Specifically, we decompose the weights of PINN's hidden layers using SVD and introduce an innovative frequency embedding hypernetwork (FEH) that links input frequencies with the singular values, enabling efficient and frequency-adaptive wavefield representation. Meta-learning is employed to provide robust initialization, improving optimization stability and reducing training time. Additionally, we implement adaptive rank reduction and FEH pruning during the meta-testing phase to further enhance efficiency. Numerical experiments, which are presented on multi-frequency scattered wavefields for different velocity models, demonstrate that Meta-LRPINN achieves much fast convergence speed and much high accuracy compared to baseline methods such as Meta-PINN and vanilla PINN. Also, the proposed framework shows strong generalization to out-of-distribution frequencies while maintaining computational efficiency. These results highlight the potential of our Meta-LRPINN for scalable and adaptable seismic wavefield modeling.
\end{abstract}

\section{\textbf{Introduction}}
Accurate and efficient wavefield modeling is essential in seismology, with applications ranging from exploration to earthquake hazard analysis \citep{carcione2002seismic, fichtner2010full}. Among various modeling techniques, frequency-domain wavefield simulation has gained prominence due to its ability to handle steady-state wave propagation across multiple frequencies \citep{marfurt1984accuracy}. By focusing on specific frequency components, frequency-domain methods not only reduce the dimensionality of the problem, but also facilitate multi-frequency analysis, which is critical for multi-scale inversion strategies \citep{pratt1999seismicpart1, pratt1999seismicpart2}. These capabilities have positioned frequency-domain modeling as a powerful approach to better understand subsurface properties \citep{pratt1990frequency}. 

Despite its advantages, frequency-domain wavefield modeling faces significant challenges in practical applications, particularly in heterogeneous media. Solving the frequency-domain wave equation requires handling large and sparse linear systems for each frequency \citep{marfurt1984accuracy, pratt1999seismicpart1}, where the impedance matrix relates wavefield solutions to seismic sources. Traditional approaches rely on numerical techniques such as finite difference and finite element methods to discretize the model and solve these systems using either direct solvers \citep{operto20073d} or iterative methods \citep{riyanti2006new, plessix2009three, petrov20123d, wu2018efficient}. Direct solvers, though precise, are computationally expensive for large-scale models, for example 3D models, due to their reliance on matrix factorization \citep{plessix2009three}. On the other hand, iterative solvers offer a more memory-efficient alternative, but often struggle with convergence, particularly in complex and high-frequency scenarios \citep{yang2004optimal, huang2021finite}. Recent advancements in preconditioning techniques have partially alleviated these computational challenges \citep{sheikh2016accelerating, belonosov2017iterative}, yet the overall cost and complexity of traditional methods remain a bottleneck for large-scale applications. 



To address the limitations of traditional numerical methods, physics-informed neural networks (PINNs) \citep{raissi2019physics} have emerged as a promising alternative for wavefield modeling \citep{rasht2022physics, waheed2022kronecker, sandhu2023multi, schuster2024review, chai2024modeling}. By embedding the governing partial differential equations (PDEs) into the training process of neural networks (NNs), PINNs eliminate the need for explicit grid discretization or labeled data. Instead, they leverage automatic differentiation to enforce physical constraints, enabling grid-free modeling and flexibility in handling complex wavefield scenarios. PINNs have demonstrated their ability to solve PDEs in various domains, offering a versatile framework for wavefield simulation. For example, \cite{song2021solving} proposed using PINNs to represent frequency-domain scattering wavefields in transversely isotropic media with a vertical symmetry axis. To further improve the performance of PINNs in wavefield representation, they also introduced an adaptive sinusoidal activation function to optimize the training process of PINNs \citep{song2022versatile}. \cite{bin2021pinneik} applied PINNs to solve the Eikonal equation for traveltime computation, which can better support tomographic imaging. \cite{huang2021modified} proposed using positional encoding to transform spatial coordinates into a higher-dimensional space, thereby accelerating the training process of PINNs and enhancing the accuracy of wavefield representation. \cite{wu2023helmholtz} incorporated perfectly matched layer (PML) boundary conditions into the loss function of PINNs and replaced the affine functions with quadratic functions to overcome challenges in representing wavefields for non-smooth velocity models. \cite{alkhalifah2024physics} used adaptive Gabor layers to enhance the fully connected NN model of vanilla PINNs, thereby improving training efficiency and the ability to represent high-frequency wavefields.

Despite these advantages, PINN-based methods still face significant challenges when applied to multi-frequency wavefields and diverse velocity models. A fundamental limitation lies in the fact that PINN is essentially a function approximator \citep{huang2022meta}. The network is trained to represent the wavefield solution for a specific velocity model and frequency configuration based on the given training data. However, when the frequency or velocity changes, the underlying physical relationships governing the wavefield also change. As a result, the trained PINN is no longer valid, and the network should be retrained from scratch to adapt to the new conditions. This retraining process is computationally expensive and undermines the scalability of PINN-based methods for practical seismic applications involving a wide range of frequencies and velocity models. 

Some prior studies have attempted to address this issue. For instance, \cite{huang2022single} proposed a reference frequency loss, utilizing the linear relationship between frequency and wavenumber to train a network capable of representing scattering wavefields with multiple frequency components simultaneously. \cite{song2023simulating} introduced Fourier-featured PINNs, which enhance the representation of multi-frequency wavefields by incorporating frequency information into the input and using positional encoding to map the input data to a higher-dimensional space. However, both approaches face the same limitation: they are only effective for the specific training velocity models. For new velocity models, training still needs to start from scratch. To improve the adaptability of PINNs to different velocities, \cite{taufik2024multiple} proposed LatentPINN, which uses an autoencoder trained across various velocity models to extract a latent representation of the velocity. The extracted latent representation, combined with spatial coordinates, serves as the input to PINNs, allowing the network to represent wavefields corresponding to different velocities. However, this approach requires aligning the size of the latent representation with the input spatial coordinates, making it applicable only to specific-sized velocity models once the autoencoder is trained. Moreover, LatentPINN may face generalization issues for out-of-distribution velocity models. In our previous work, we proposed Meta-PINN \citep{cheng2025meta}, which uses a meta-learning algorithm \citep{finn2017model} to learn a robust initialization on a very limited set of velocity models, allowing PINNs to converge quickly to an ideal accuracy on new velocity models. However, both \cite{taufik2024multiple}'s work and our previous work focused only on single-frequency wavefields, and they still require training from scratch for out-of-distribution frequency wavefields. 

Here, we extend our previous Meta-PINN to provide a generalized framework capable of representing multi-frequency wavefields for varying velocity models. We call our more general framework as Meta-LRPINN, where we integrate low-rank decomposition and frequency embedding to enhance efficiency and adaptability. By applying singular value decomposition (SVD) to the weight matrices of the PINN, we can significantly reduce the number of parameters, enabling efficient representation of complex wavefields. Meanwhile, we develop a frequency embedding hypernetwork (FEH) to embed frequency information directly into the singular values, allowing the network to dynamically adapt to multi-frequency wavefields. To further improve initialization and scalability, we employ a meta-learning strategy that optimizes the initialization parameters for both the LRPINN and the FEH. During the meta-training stage, the framework learns generalizable initialization parameters across various velocity models and frequencies, enabling rapid convergence during meta-testing phase with only a very small number of gradient updates. Moreover, by pruning the FEH in the meta-testing stage, we reduce computational complexity while retaining the frequency-aware capabilities of the network. Furthermore, to address the representation requirements for different frequency components while considering computational resource limitations, we propose an adaptive rank reduction strategy to further enhance training efficiency and reduce memory consumption, while still achieving competitive wavefield representations.

The proposed Meta-LRPINN framework provides a scalable, efficient, and adaptive solution for frequency-domain wavefield modeling, addressing key limitations of existing PINN methods. Our contributions are summarized as follows:
\begin{itemize}
    \item We propose a Meta-LRPINN framework that leverages singular value decomposition and frequency embedding to efficiently model multi-frequency seismic wavefields.
    \item We develop a meta-learning-based initialization strategy to enhance the scalability and adaptability of Meta-LRPINN across different velocity models and frequencies.
    \item We propose a rank-adaptive reduction strategy to accommodate the representation of different frequency wavefields while addressing computational resource constraints.
    \item We validate the proposed framework through extensive numerical examples, demonstrating its effectiveness for various frequencies and complex velocity scenarios. 
\end{itemize}

\section{\textbf{Reivew of frequency-domain scattered wavefield solutions with vanilla PINN}}\label{method_review}
Modeling seismic wave propagation in acoustic, isotropic media with constant density involves solving the frequency-domain acoustic wave equation, which is given by \citep{Aki1980QuantitativeST}: 
\begin{equation}\label{eq1} 
\omega^2 m(\mathbf{x}) u (\mathbf{x},\mathbf{x}_s,\omega) +  
\nabla^2 u (\mathbf{x},\mathbf{x}_s,\omega) = s(\mathbf{x},\mathbf{x}_s,\omega),
\end{equation} 
where $\omega$ denotes the angular frequency, $\mathbf{x} = (x,z)$ is spatial coordinates in a 2-D medium, $\mathbf{x}_s=(x_s,z_s)$ represents the source position, $m(\mathbf{x}) = 1/v^2(\mathbf{x})$ is the spatially varying squared slowness corresponding to velocity $v(\mathbf{x})$, $u(\mathbf{x}, \mathbf{x}_s,\omega)$ is the complex wavefield in the frequency domain, $s(\mathbf{x},\mathbf{x}_s,\omega)$ represents the source term, and $\nabla^2$ is the Laplacian operator with respect to the spatial coordinates. 

Recently, PINNs have been proposed as an effective tool for solving wave equations by embedding the governing equations into the training process of NNs \citep{raissi2019physics}. PINNs enable the approximation of wavefield solutions without requiring labeled data, utilizing automatic differentiation to enforce the physical laws described by the wave equation. However, the singularity in Equation (\ref{eq1}) caused by a point representation of the source poses challenges for PINNs, as it hinder the convergence speed and reduce the accuracy of the wavefield solution. 

To overcome this issue, \cite{alkhalifah2021wavefield} proposed reformulating the equation using a scattered wavefield approach based on scattering theory. This leads to the following perturbation equation:
\begin{equation}\label{eq2} 
\omega^2 m(\mathbf{x}) \delta u(\mathbf{x},\mathbf{x}_s,\omega) + \nabla^2 \delta u(\mathbf{x},\mathbf{x}_s,\omega) = -\omega^2 \delta m(\mathbf{x}) u_0(\mathbf{x},\mathbf{x}_s,\omega), 
\end{equation}
where $\delta u(\mathbf{x},\mathbf{x}_s,\omega) = u(\mathbf{x},\mathbf{x}_s,\omega) - u_0(\mathbf{x},\mathbf{x}_s,\omega)$ is the scattered wavefield, $u_0(\mathbf{x},\mathbf{x}_s,\omega)$ is the known background wavefield corresponding to a homogeneous medium with constant velocity $v_0$, and $\delta m(\mathbf{x}) = m(\mathbf{x}) - m_0$ is the slowness perturbation with $m_0 = 1/v_0^2$. 

In a homogeneous and infinite background medium, the background wavefield $u_0(\mathbf{x}, \mathbf{x}_s,\omega)$ can be analytically computed using:
\begin{equation}\label{eq3} 
u_0(\mathbf{x},\mathbf{x}_s,\omega) = \frac{\mathrm{i}}{4} H_0^{(2)}\left( \frac{\omega}{v_0} \left| \mathbf{x} - \mathbf{x}_s \right| \right), 
\end{equation}
where $H_0^{(2)}$ denotes the zero-order Hankel function of the second kind, $\mathrm{i}$ is the imaginary unit, and $\left| \cdot \right|$ represents the Euclidean distance. This equation allows us to quickly obtain the analytical wavefield solution at arbitrary spatial positions. 

By leveraging this reformulation, \cite{song2021solving} employed a vanilla PINN to directly predict the scattered wavefield $\delta u(\mathbf{x},\mathbf{x}_s,\omega)$, where the inputs are the spatial coordinates $\mathbf{x}$ and source position $\mathbf{x}_s$, and the outputs are the real part $\delta u_R(\mathbf{x},\mathbf{x}_s,\omega)$ and the imaginary part $\delta u_I(\mathbf{x},\mathbf{x}_s,\omega)$ of the scattered wavefield. The vanilla PINN is constructed as a fully connected NN, which can be represented by the mapping:
\begin{equation}\label{eq4} 
\delta u(\mathbf{x},\mathbf{x}_s,\omega) = \delta u_R(\mathbf{x},\mathbf{x}_s,\omega) + \mathrm{i} \delta u_I(\mathbf{x},\mathbf{x}_s,\omega) = \text{NN}_\theta(\mathbf{x},\mathbf{x}_s,\omega), 
\end{equation}
where $\text{NN}_\theta(\mathbf{x},\mathbf{x}_s,\omega)$ represents the PINN parameterized by weights and biases $\boldsymbol{\theta}$.  

The architecture of a vanilla PINN includes multiple fully connected hidden layers, which can be represented by:
\begin{equation}\label{eq5} 
\begin{aligned} 
h^{(1)} &= f[W_l \cdot (\mathbf{x}, \mathbf{x}_s) + b_l], \quad l = 1, \\
h^{(l)} &= f[W_l \cdot h^{(l)} + b_l], \quad l = 2, \ldots, L-1, \\
h^{(L)} &= W_L \cdot h^{(L)} + b_L, 
\end{aligned} 
\end{equation}
where $h^{(l)}$ represents the activations in layer $l$, $W_l$ and $b_l$ are the weights and biases of layer $l$, and $f[\cdot]$ is the activation function, often chosen to be a sine or another nonlinear function to introduce nonlinearity. The output layer directly provides the real part $\delta u_R(\mathbf{x},\mathbf{x}_s,\omega)$ and the imaginary part $\delta u_I(\mathbf{x},\mathbf{x}_s,\omega)$ of the scattered wavefield, without applying any activation function (linear layer). 

The training of the PINN is guided by a physics-informed loss function:
\begin{equation}\label{eq6}
\mathcal{L}_{p} = \frac{1}{N}\sum\limits_{j=1}^{N} \left ( {\left|(\omega^2 m^j + \nabla^2)\delta u_R^j + \omega^2 \delta m^j u_{R0}^j \right|_2^2 } \right. 
\left. {+ \left|(\omega^2 m^j + \nabla^2)\delta u_I^j + \omega^2 \delta m^j u_{I0}^j \right|_2^2 }\right),
\end{equation}
where $N$ is the number of training points sampled from the spatial domain, and $u_{R0}$ and $u_{I0}$ are the real and imaginary parts of the background wavefield $u_0(\mathbf{x}, \mathbf{x}_s, \omega)$, respectively. Since we use the network to approximate the nonlinear relationship from the input spatial position to the scattered wavefield values at the input spatial position, we can use automatic differentiation to compute the required spatial derivatives in the loss function. This provides a key advantage where we eliminate the need for grid-based discretization, allowing the network to learn a solution that inherently satisfies the differential operators across the entire domain. 

However, when we represent the scattered wavefield for large-scale, complex velocity models or with high-frequency, it typically requires deeper and wider NN architectures \citep{alkhalifah2021wavefield}. This is because large-scale complex velocity models exhibit significant spatial heterogeneity and discontinuities, which imposes higher demands on the network's representational capacity. High-frequency wavefields contain intricate details and rapid variations that shallow networks struggle to effectively capture. However, larger networks require storing a substantial number of parameters, and during training, the activation values and gradients at each layer consume significant memory resources. In addition, increasing the network's depth and width, which leads to a greater number of parameters, makes the optimization process more complex by not only adding to the computational burden but also increasing the risk for gradient vanishing or explosion, resulting in inefficient training and difficulty in convergence. 

On the other hand, since PINNs are fundamentally designed to approximate functions based on a PDE loss corresponding to specific parameters (like velocity model, frequency or source location), they face limitations when generalizing to new parameters. For instance, if the velocity model or the frequency changes, the previously trained PINN may fail to accurately predict the wavefield for these new conditions. The lack of generalization stems from the fact that the network representation is tightly coupled to the specific training velocity model or frequency. This necessitates retraining for each new scenario, which is computationally inefficient. 

To address these challenges, in the following section, we employ singular value decomposition (SVD) to represent the network with fewer, more effective parameters, thereby significantly addressing the burden of representing high-frequency wavefields and large-scale velocity models on network size. We, also, utilize a meta-training framework to improve the generalization capabilities of the PINN. Our method specifically aims to overcome the issues related to adapting the network to new velocity models and different frequencies, by training the network from a learned initialization, thereby significantly improving efficiency and scalability. 

\section{\textbf{Method}}\label{method}
\subsection{Low-rank decomposition of the network parameters}
In the context of machine learning, weight matrices in NNs often exhibit significant redundancy. This means that they can be effectively approximated by lower-rank representations without a notable loss in accuracy. This property is particularly advantageous given the growing size of modern NNs, as reducing the dimensionality of weight matrices can yield substantial savings in memory and computational cost, while also improving generalization by mitigating overfitting. 

To address the redundancy, we consider a low-rank decomposition of the weight matrices $W_l$ of each layer $l$ in the PINN, the decomposition can be expressed as:
\begin{equation}\label{eq7}
W_l = U_l \cdot V_l^\text{T}, \quad l = 1, \ldots, L,
\end{equation}
where $U_l \in \mathbb{R}^{m \times k}$ and $V_l \in \mathbb{R}^{n \times k}$, with $k \ll \min(m, n)$. Here, $k$ denotes the rank of the approximation. This factorization significantly reduces the number of parameters because the combined size of $U_l$ and $V_l$ is much smaller than the original size of $W_l$, for small k. Consequently, this decomposition reduces the computational burden and improves the network’s efficiency, enabling faster training and inference. 

While the conventional low-rank decomposition reduces the number of parameters, it lacks the structured interpretability required for modeling frequency-dependent wavefields. To address this limitation, we further extend it to a more informative low-rank decomposition, Singular Value Decomposition (SVD). SVD factorizes the weight matrix $W_l$ into three components:
\begin{equation}\label{eq8}
W_l = U_l \cdot \Theta_l \cdot V_l^\text{T},
\end{equation}
where $\Theta_l$ is a diagonal matrix containing singular values:
\begin{equation}\label{eq9}
\Theta_l = \text{diag}(\sigma_{l,1}, \sigma_{l,2}, \dots, \sigma_{l,k}) = 
\begin{bmatrix}
\sigma_{l,1} & 0 & \cdots & 0 \\
0 & \sigma_{l,2} & \cdots & 0 \\
\vdots & \vdots & \ddots & \vdots \\
0 & 0 & \cdots & \sigma_{l,k}
\end{bmatrix}.\\
\end{equation}

Compared to the conventional low-rank decomposition, SVD introduces an additional diagonal matrix $\Theta_l$, which contains only $k$ singular values. The slight increase in parameter count is negligible relative to the memory and computational savings achieved by the decomposition. 

Despite the small increase in parameters, SVD brings significant advantages in terms of representation and interpretability. First, it provides a more structured decomposition that separates the weight matrix into orthonormal bases ($U_l$ and $V_l$) and singular values ($\Theta_l$). This structured factorization separates the geometric transformations (captured by \(U_l\) and \(V_l\)) from their scaling (captured by \(\Theta_l\)), enabling a more interpretable representation of the network's internal operations. As a result, it provides a more detailed understanding of how different components of the network contribute to the overall transformation. 

Moreover, the singular values, $\sigma_{l,i}$, $i=1,\ldots,k$, dominate the contribution of each rank-1 component to the transformation performed by the layer. By associating these singular values with frequency-dependent information, we can directly encode and control the frequency components of the scattered wavefield within the network. This is particularly important as the scattered wavefield inherently varies with frequency. Thus, by applying SVD to the weight matrices of the PINN, we can not only reduce the parameter count but also achieve a direct link between the network's internal representation and the frequency content of the wavefield. This enhanced representation improves the generalization capability of the model, particularly for new frequency and velocity scenarios, without requiring retraining from scratch. \\

In the following section, we will introduce how to explicitly link the frequency of scattered wavefield to the singular values of the weight matrix. This approach allows us to incorporate frequency information more directly into the network, thereby enhancing its adaptability and efficiency in simulating scattered wavefields at different frequencies. 

\begin{figure}[htbp]
\centering
\includegraphics[width=0.98\textwidth]{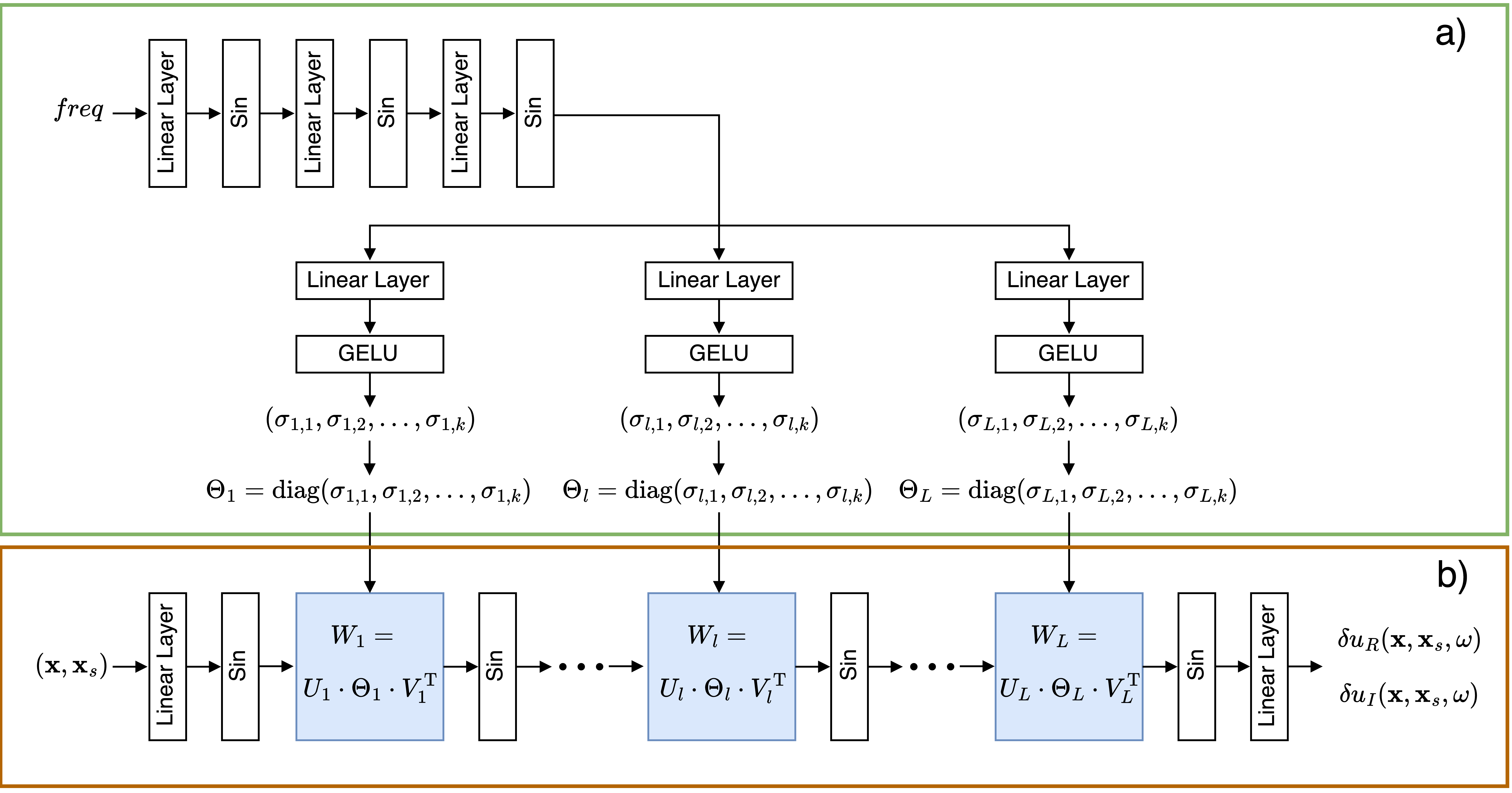}
\caption{
Illustration of the proposed LRPINN with frequency embedding. 
(a) \textbf{Frequency embedding hypernetwork}: This network maps the input frequency (\(\textit{freq}\)) to frequency-dependent singular values (\(\sigma_{l,1}, \sigma_{l,2}, \dots, \sigma_{l,k}\)) for each layer through a series of linear layers with sine and GELU activations. These singular values are used to construct diagonal matrices (\(\Theta_l\)) that dynamically adapt to the input frequency. 
(b) \textbf{Low-rank PINN}: The inputs (spatial location $\mathbf{x} =(x,z)$ and source location $\mathbf{x}_s =(x_s,z_s)$) are encoded using sine activations, and each layer's weight matrix (\(W_l\)) is factorized into \(U_l \cdot \Theta_l \cdot V_l^\text{T}\), incorporating the frequency-dependent singular values from (a). The network outputs the real (\(\delta u_R(\mathbf{x}, \mathbf{x}_s, \omega)\)) and imaginary (\(\delta u_I(\mathbf{x}, \mathbf{x}_s, \omega)\)) components of the scattered wavefield.}
\label{fig1}
\end{figure}

\subsection{Low-rank PINN with frequency embedding}
To address the challenges of modeling scattered wavefields across a wide range of frequencies, we propose an innovative low-rank PINN (LRPINN) architecture that incorporates frequency embedding. By explicitly linking the frequency of the scattered wavefield to the singular values of each layer's weight matrix, the proposed framework achieves enhanced adaptability and scalability for different frequency scenarios. 

The architecture is illustrated in Figure~\ref{fig1}, which consists of two main components:
\begin{enumerate}
    \item \textbf{Frequency embedding hypernetwork (FEH)} (Figure \ref{fig1}a): \\
    The FEH is designed to link the input frequency $\textit{freq}$ to the singular values of each layer’s weight matrix. Specifically, the FEH begins by feeding the frequency into a series of linear layers followed by sine activation functions to extract frequency-specific features. This helps the network capture the complex relationships between frequency and the wavefield characteristics. The output from these linear layers with sine activation is then processed through additional linear layers followed by gaussian error linear unit (GELU) activation functions, resulting in the singular values for each hidden layer. For each layer $l$, the network produces a vector $\sigma_l=(\sigma_{l,1}, \sigma_{l,2}, \dots, \sigma_{l,k})$, which is then used to construct the diagonal singular value matrix: 
    \begin{equation}
    \Theta_l = \text{diag}(\sigma_{l,1}, \sigma_{l,2}, \dots, \sigma_{l,k}).
     \end{equation}
    The singular values produced by the FEH, denoted by $\Theta_l$, are frequency-dependent and they dynamically update the low-rank decomposition of each hidden layer’s weight matrix in the LRPINN. 
    \item \textbf{LRPINN with frequency-aware weights} (Figure \ref{fig1}b): \\
    The second component of the proposed architecture is the LRPINN, which represents the scattered wavefield by utilizing low-rank decomposed weights that are modulated based on the input frequency. The spatial coordinates $(\mathbf{x}, \mathbf{x}_s)$ are first passed through a linear layer followed by sine activation functions to encode spatial features effectively. Each hidden layer of the LRPINN is represented by a weight matrix $W_l$, which is factorized using SVD as shown in Equation \ref{eq8}. Here, $U_l \in \mathbb{R}^{m \times k}$ and $V_l \in \mathbb{R}^{n \times k}$ are learned matrices, and $\Theta_l$ is the frequency-dependent diagonal matrix of singular values produced by the FEH. We emphasize that, for further simplification, we here discard the bias of each hidden layer. In other words, the input of each hidden layer is directly multiplied by the constructed $W_l$ without adding the bias, but is directly activated. After passing through the sequence of layers, the network produces the real and imaginary components of the scattered wavefield, $\delta u_R(\mathbf{x}, \mathbf{x}_s, \omega)$ and $\delta u_I(\mathbf{x}, \mathbf{x}_s, \omega)$, respectively, which represent the final predictions of our LRPINN.

\end{enumerate}

By embedding the frequency information into the singular values of each layer, our proposed architecture allows the weight matrix $W_l$ to change based on the frequency of the input, effectively adapting the representation of each hidden layer to match the characteristics of the given frequency. Meanwhile, the low-rank representation reduces the number of parameters needed to represent the weight matrices and, thus, improves the network's ability to represent the scattered wavefield across different frequencies effectively and efficiently. Furthermore, the frequency embedding provides an efficient way to adjust these parameters based on the frequency of the input data. For example, we can adaptively adjust the network weights by removing smaller singular values and their corresponding matrices ($\delta u_R(\mathbf{x}, \mathbf{x}_s, \omega)$ and $\delta u_I(\mathbf{x}, \mathbf{x}_s, \omega)$) based on their magnitudes, as we will see later. This strategy helps make the network more compact and computationally efficient.

\subsection{Meta-learning enhanced LRPINN}
While the proposed LRPINN with frequency embedding demonstrates significant improvements in efficiency and adaptability, its performance is highly sensitive to the quality of the initial weight matrices. Random initialization, which is commonly used, often results in slow convergence and suboptimal solutions. This issue is particularly pronounced in low-rank settings, as the reduced parameterization limits the network's capacity to explore the solution space effectively during optimization. Consequently, achieving optimal performance with the LRPINN can be challenging without a well-designed initialization strategy. 

To address this limitation, we, in this subsection, propose using meta-learning \citep{finn2017model} to provide a robust initialization for the LRPINN, where we call this framework as Meta-LRPINN. Meta-LRPINN consists of two key stages: meta-training and meta-testing, as outlined in Algorithm \ref{alg1} and Algorithm \ref{alg2}, respectively. 

The goal of the meta-training stage is to find an optimal initialization for LRPINN parameters $\boldsymbol{\theta}$ and FEH parameters $\boldsymbol{\epsilon}$ that allows for fast adaptation across a range of velocity $v(\mathcal{T})$ and frequency $\textit{freq}(\mathcal{T})$ distributions. Here, $mathcal{T}$ represents all training data set, which can be considered as all velocity models and their corresponding frequencies. This stage involves a double-loop optimization strategy, consisting of an inner loop for task-specific adaptation and an outer loop for meta-optimization. 

The inner loop begins by copying the initialization parameters, which is denoted by ($\boldsymbol{\theta}_i', \boldsymbol{\epsilon}_i'$), from the previous iteration. These parameters are then updated to adapt specifically to a sampled support dataset ($\mathcal{T}_i^s$) and corresponding frequency ($\textit{freq}^s$), i.e., a velocity model and the corresponding frequency sampled from the training data set. The task-specific loss $\mathcal{L}_{T_i^s, \textit{freq}^s}$ is computed based on the support dataset, and gradient descent is applied to adjust the parameters:
\begin{equation}\label{eq11}
(\boldsymbol{\theta}_i', \boldsymbol{\epsilon}_i') = (\boldsymbol{\theta}_i', \boldsymbol{\epsilon}_i') - lr_{\text{inner}} \cdot \nabla_{\boldsymbol{\theta}_i', \boldsymbol{\epsilon}_i'} \mathcal{L}_{\mathcal{T}_i^s, \textit{freq}^s}(G_{\boldsymbol{\theta}_i', \boldsymbol{\epsilon}_i'}),
\end{equation}
where $lr_{\text{inner}}$ is the inner Loop learning rate, and $G_{\boldsymbol{\theta}_i', \boldsymbol{\epsilon}_i'}$ represents a parameterized function with the copied parameters ($\boldsymbol{\theta}_i', \boldsymbol{\epsilon}_i'$). The updated parameters $(\boldsymbol{\theta}_i', \boldsymbol{\epsilon}_i')$ are specific to the current support dataset and the corresponding frequency, representing the model's adaptation to the given velocity and frequency. 

In the outer loop, the adapted parameters from the inner loop $(\boldsymbol{\theta}_i', \boldsymbol{\epsilon}_i')$ are directly evaluated on the corresponding query dataset $T_i^q$ and the corresponding frequency $\textit{freq}^q$. The query dataset loss $\mathcal{L}_{\mathcal{T}_i^q, \textit{freq}^q}$ measures how well the task-specific adaptation generalizes to new velocity and frequency. This loss is computed for all sampled support-query pairs, and the total query loss is accumulated:
\begin{equation}\label{eq12}
\mathcal{L}_{\text{sum}} = \sum_i \mathcal{L}_{\mathcal{T}_i^q, \textit{freq}^q}(G_{\boldsymbol{\theta}_i', \boldsymbol{\epsilon}_i'}).
\end{equation}
The accumulated loss is then used to compute the gradients with respect to the global initialization parameters ($\boldsymbol{\theta}, \boldsymbol{\epsilon}$), which are updated using gradient descent:
\begin{equation}\label{eq13}
(\boldsymbol{\theta}, \boldsymbol{\epsilon}) \leftarrow (\boldsymbol{\theta}, \boldsymbol{\epsilon}) - lr_{\text{meta}} \cdot \nabla_{\boldsymbol{\theta}, \boldsymbol{\epsilon}} \mathcal{L}_{\text{sum}},
\end{equation}
where $lr_{\text{meta}}$ is the outer loop learning rate. This update ensures that the initialization parameters ($\boldsymbol{\theta}, \boldsymbol{\epsilon}$) are optimized to perform well across all tasks (velocity models and corresponding frequencies). 

The inner loop and outer loop are repeated for multiple iterations. In each iteration, we sample $N$ pairs of support-query datasets along with the corresponding $(\textit{freq}^s, \textit{freq}^q)$ pairs from the given velocity and frequency distributions, which can be considered as $N$ tasks. Each iteration will then iterate over these $N$ pairs of datasets. This iterative process allows the model to learn a robust initialization that generalizes across tasks, enabling effective task-specific adaptation during the meta-testing stage. 

In the meta-testing stage, the meta-trained initialization is used to adapt the LRPINN to specific velocity models $v$ and frequency $\textit{freq}$, leveraging the knowledge learned during meta-training phase. Specifically, first, the FEH, which accept the test frequency $\textit{freq}$, predicts the singular values for each layer of the LRPINN: 
\begin{equation} \label{eq14}
\sigma_l = (\sigma_{l,1}, \sigma_{l,2}, \dots, \sigma_{l,k}) = \text{FEH}(\textit{freq}). 
\end{equation} 
After this step, the FEH is pruned to reduce computational complexity and, also, further improve computational efficiency, leaving only the LRPINN fine-tuned for the specific task. Next, the orthogonal matrices $(U_l, V_l)$ are extracted from the LRPINN’s meta-initialized weights and combined with the predicted singular values $\sigma_l$ to construct the weight matrices dynamically: 
\begin{equation}\label{eq15}
W_l = U_l \cdot \Theta_l \cdot V_l^\text{T}, 
\end{equation} 
where $\Theta_l = \text{diag}(\sigma_{l,1}, \sigma_{l,2}, \dots, \sigma_{l,k})$, and the orthogonal matrices $U_l$ and $V_l$ and the singular values $\sigma_l$ are set as learnable parameters. These learnable parameters $(U_l, V_l, \sigma_l)$ are then fine-tuned using gradient descent: 
\begin{equation}\label{eq16}
(U_l, V_l, \sigma_l) \leftarrow (U_l, V_l, \sigma_l) - lr \cdot \nabla_{U_l, V_l, \sigma_l} \mathcal{L}_{v, \textit{freq}}F_{U_l, V_l, \sigma_l}, 
\end{equation} 
where $lr$ is the meta-testing learning rate, and $F_{U_l, V_l, \sigma_l}$ is a parameterized function to represent the LRPINN with the learnable parameters $(U_l, V_l, \sigma_l)$. 

Compared to our previous work, Meta-PINN \citep{cheng2025meta}, the Meta-LRPINN presented here leverages not only the prior knowledge of the velocity distribution, but also the frequency distribution to learn an optimized initialization, in our new SVD based representation of PINN weights. This initialization allows the network to achieve faster convergence and better performance across new velocity models and various frequencies. The SVD representation, on the other hand, allows for a low-rank representation of the PINN network weights, which effectively enhances computational efficiency and reduces memory consumption. Furthermore, Meta-LRPINN can refine the low-rank representation by pruning less significant components while simplifying the model by removing the FEH. This approach maintains the efficiency benefits of the LRPINN while further improving its adaptability and generalization capabilities. 

\begin{algorithm}
\caption{Meta-LRPINN: Meta-Training}\label{alg1}
\textbf{Input:} ${v(\mathcal{T})}$: The full velocity dataset. \\
\textbf{Input:} ${\textit{freq}(\mathcal{T})}$: The full frequency distribution. \\
\textbf{Input:} ${lr_{inner}, lr_{meta}}$: Learning rates for inner and outer loops, respectively. \\
\textbf{Input:} ${iter}$: The number of iterations in the support dataset. \\
\textbf{-------------------------------- Meta-training stage -----------------------------} \\
\textbf{Output:} Meta-based initialization of the LRPINN and FEH
\begin{algorithmic}
\State 1: Randomly initialize LRPINN parameters $\boldsymbol{\theta}$ and FEH parameters $\boldsymbol{\epsilon}$
\State 2: \textbf{while} all velocity model ${v(\mathcal{T})}$ and frequencies ${\textit{freq}(\mathcal{T})}$ \textbf{do}
\State 3: \quad Sample batches of support $\mathcal{T}_i^s$ and query $\mathcal{T}_i^q$ datasets from the full dataset $v(\mathcal{T})$
\State 4: \quad Sample the corresponding frequencies $\textit{freq}^s$ and $\textit{freq}^q$ from ${\textit{freq}(\mathcal{T})}$ to \\
\quad \quad \quad \quad \quad \quad the sampled batches of support and query datasets, respectively
\State 5: \quad \textbf{for} every $(\mathcal{T}_i^s, \textit{freq}^s)$ and $(\mathcal{T}_i^q, \textit{freq}^q)$ \textbf{do}
\State 6: \quad \quad Copy the Meta-LRPINN model $G_{\boldsymbol{\theta}^{'}, \boldsymbol{\epsilon}^{'}} = G_{\boldsymbol{\theta}, \boldsymbol{\epsilon}}$
\State 7: \quad \quad \textbf{for} ${i}$ \textbf{in} ${iter}$ \textbf{do}
\State 8: \quad \quad \quad \quad Evaluate $\nabla_{\boldsymbol{\theta}^{'}, \boldsymbol{\epsilon}^{'}} \mathcal{L}_{\mathcal{T}_i^s, \textit{freq}^s} \left( G_{\boldsymbol{\theta}^{'}, \boldsymbol{\epsilon}^{'}} \right)$ with respect to \\
\quad \quad \quad \quad \quad \quad \quad \quad \quad the sampled support dataset $\mathcal{T}_i^s$ and frequency $\textit{freq}^s$
\State 9: \quad \quad \quad \quad Compute adapted parameters with gradient descent: \\
\quad \quad \quad \quad \quad \quad \quad \quad \quad $(\boldsymbol{\theta}_i^{'}, \boldsymbol{\epsilon}_i^{'})  = (\boldsymbol{\theta}_i^{'}, \boldsymbol{\epsilon}_i^{'}) - lr_{inner} \cdot \nabla_{\boldsymbol{\theta}_i^{'}, \boldsymbol{\epsilon}_i^{'}} \mathcal{L}_{\mathcal{T}_i^s, \textit{freq}^s} \left( G_{\boldsymbol{\theta}_i^{'}, \boldsymbol{\epsilon}_i^{'}} \right)$
\State 10: \quad \quad \textbf{end for}
\State 11: \quad \quad Evaluate $ \mathcal{L}_{\mathcal{T}_i^q, \textit{freq}^q}( G_{\boldsymbol{\theta}_i^{'}, \boldsymbol{\epsilon}_i^{'}})$ with respect to the sampled query dataset $\mathcal{T}_i^q$ and frequency $\textit{freq}^q$
\State 12: \quad \textbf{end for}
\State 13: \quad Sum the loss of all samples on the query dataset $\mathcal{T}_i^q$ and the corresponding frequencies $\textit{freq}^q$: \\
\quad \quad \quad \quad \quad \quad \quad \quad $\mathcal{L}_{sum} = \sum_{\mathcal{T}_i^q, \textit{freq}^q} \mathcal{L}_{\mathcal{T}_i^q, \textit{freq}^q} ( G_{{\boldsymbol{\theta}}_i^{'}, {\boldsymbol{\epsilon}}_i^{'}})$
\State 14: \quad Update the Meta-LRPINN $(\boldsymbol{\theta}, \boldsymbol{\epsilon}) \leftarrow (\boldsymbol{\theta}, \boldsymbol{\epsilon}) - lr_{meta} \cdot \nabla_{\boldsymbol{\theta}, \boldsymbol{\epsilon}} \mathcal{L}_{sum}$
\State 15: \textbf{end while}
\State 16: \textbf{Return:} Meta-LRPINN parameters $(\boldsymbol{\theta}, \boldsymbol{\epsilon})$
\end{algorithmic}
\end{algorithm}

\begin{algorithm}
\caption{Meta-LRPINN: Meta-Testing}\label{alg2}
\textbf{Input:} $v$: Test velocity model. \\
\textbf{Input:} $\textit{freq}$: Test frequency of the scattered wavefield. \\
\textbf{Input:} $lr$: Learning rate for meta-testing stage. \\
\textbf{Input:} ${iter}$: The number of iterations in the meta-testing stage. \\
\textbf{-------------------------------- Meta-testing stage -----------------------------} \\
\textbf{Output:} Velocity- and frequency-specific Meta-LRPINN model
\begin{algorithmic}
\State 1: Load Meta-LRPINN initialization parameters $(\boldsymbol{\theta}, \boldsymbol{\epsilon})$ from Meta-training stage
\State 2: Input frequency $\textit{freq}$ to FEH to obtain \\
\quad \quad \quad \quad the singular values of each layer $l$ of LRPINN: $(\sigma_{l,1}, \sigma_{l,2}, \dots, \sigma_{l,k}) = \text{FEH}(\textit{freq})$
\State 3: Turn predicted singular values $\sigma_{l} = (\sigma_{l,1}, \sigma_{l,2}, \dots, \sigma_{l,k})$ into learnable parameters and prune FEH
\State 4: Extract the matrices $U_l \in \mathbb{R}^{m \times k}$ and $V_l \in \mathbb{R}^{n \times k}$ from Meta-LRPINN initialization parameters $\boldsymbol{\theta}$ 
\State 5: Set matrices $U_l \in \mathbb{R}^{m \times k}$ and $V_l \in \mathbb{R}^{n \times k}$ to be learnable
\State 6: \textbf{for} ${epoch}$ \textbf{in} ${iter}$ \textbf{do}
\State 7: \quad \quad Construct diagonal matrices $\Theta_l$ from the learnable singular values $\sigma_{l}$
\State 8: \quad \quad Calculate the weight matrix $W_l$ of each layer of LRPINN: $W_l = U_l \cdot \Theta_l \cdot V_l^\text{T}$
\State 9: \quad \quad Evaluate $\nabla_{U_l, V_l, \sigma_{l}} \mathcal{L}_{v, \textit{freq}} \left( F_{U_l, V_l, \sigma_{l}} \right)$ with respect to \\
\quad \quad \quad \quad \quad \quad the test velocity model and the specific frequency
\State 10: \quad \quad Update the learnable parameters with gradient descent: \\
\quad \quad \quad \quad \quad \quad \quad \quad \quad $(U_l, V_l, \sigma_{l}) = (U_l, V_l, \sigma_{l}) - lr \cdot \nabla_{U_l, V_l, \sigma_{l}} \mathcal{L}_{v, \textit{freq}} \left( F_{U_l, V_l, \sigma_{l}} \right)$
\State 11: \textbf{end for}
\State 12: \textbf{Return:} Velocity- and frequency-specific Meta-LRPINN model
\end{algorithmic}
\end{algorithm}

\subsection{Adaptive rank reduction}\label{method4}
As described in the previous subsections, we chose SVD over traditional low-rank decomposition because SVD provides unique advantages, including a more interpretable representation of the network's internal operations. In SVD, the singular values can be seen as scaling factors for the geometric transformations represented by the orthogonal matrices $U_l$ and $V_l$. This characteristic indicates that larger singular values correspond to significant contributions, while smaller singular values contribute marginally to the overall transformation. 

For seismic wavefields, larger singular values are likely associated with low-frequency components, while smaller singular values may capture high-frequency details. This insight allows us to perform adaptive rank reduction in scenarios where certain wavefield characteristics, such as low-frequency representations, are prioritized. By adaptively reducing the rank of SVD based on the magnitude of the singular values, we can further improve the computational efficiency and memory usage during the meta-testing stage. 

Specifically, during the meta-testing stage, we can first obtain a set of singular values $\sigma_l = (\sigma_{l,1}, \sigma_{l,2}, \dots, \sigma_{l,k})$ for each layer $l$ of the LRPINN by inputting the frequency ($\textit{freq}$) into the FEH, as shown in Equation \ref{eq14}. We then adaptively prune smaller singular values based on their magnitudes. Let $ \tau $ represent a threshold for pruning, such that only singular values satisfying $|\sigma_{l,i}| \geq \tau$ are retained. Mathematically, the pruned singular values can be represented as:
\begin{equation}\label{eq17}
\sigma_l' = \{\sigma_{l,i} \mid |\sigma_{l,i}| \geq \tau, \, i = 1, 2, \dots, k\},
\end{equation}
where the threshold $\tau$ can be adjusted based on task requirements, allowing for a balance between model complexity and representational capacity. Actually, in our implementation, instead of directly choosing a fixed threshold $\tau$, the threshold is determined by the desired percentage of retained ranks. For example, if we aim to retain 50\% of the rank, we sort the singular values by their absolute magnitudes and select the largest 50\%:
\begin{equation}\label{eq18}
\text{indices}(\sigma_l') = \text{top-k}(|\sigma_l|, \lfloor r \cdot k \rfloor),
\end{equation}
where $r$ is the retention ratio (e.g., $r=0.5$ for 50\%), $k$ is the total number of singular values, and $\text{top-k}$ returns the indices of the $\lfloor r \cdot k \rfloor$ largest absolute values. The retained singular values are then given by:
\begin{equation}\label{eq19}
\sigma_l' = \{\sigma_{l,i} \mid i \in \text{indices}(\sigma_l')\}.
\end{equation}

The corresponding rows and columns in the orthogonal matrices $U_l$ and $V_l$ that align with the pruned singular values are also removed:
\begin{equation}\label{eq20}
U_l' = U_l[:, \text{indices}(\sigma_l')], \quad V_l' = V_l[:, \text{indices}(\sigma_l')],
\end{equation}
where $\text{indices}(\sigma_l')$ denotes the indices of the retained singular values in $\sigma_l$. Using the pruned components, the weight matrix $W_l$ for each layer is reconstructed dynamically as:
\begin{equation}\label{eq21}
W_l' = U_l' \cdot \Theta_l' \cdot (V_l')^\text{T},
\end{equation}
where $\Theta_l' = \text{diag}(\sigma_l')$ is the diagonal matrix containing the retained singular values. 

By removing smaller singular values and their corresponding dimensions in $U_l$ and $V_l$, the number of learnable parameters is reduced, significantly decreasing computational and memory costs during meta-testing. This adaption is particularly beneficial in scenarios where low-frequency representations dominate or computational resources are limited.

\subsection{Loss functions}
For our Meta-LRPINN, in addition to the original physical and regularization losses from the Meta-PINN framework, we introduce an additional loss term to ensure the orthogonality of the matrices $U_l$ and $V_l$ in the SVD decomposition, which has the form as follows: 
\begin{equation}\label{eq22} 
\mathcal{L}_{\text{ort}} = \sum_{l=1}^L \left( |U_l^\text{T} U_l - \mathbf{I}|_F^2 + |V_l^\text{T} V_l - \mathbf{I}|_F^2 \right), 
\end{equation} 
where $L$ is the total number of layers, $\mathbf{I}$ is the identity matrix, and $|\cdot|_F$ denotes the Frobenius norm. This term strives to make $U_l$ and $V_l$ orthonormal, which is crucial for the validity of the SVD-based low-rank representation. 

Thus, the total loss combines the three components: 
\begin{equation}\label{eq23} 
\mathcal{L}_{\text{total}} = \lambda_{scale} \cdot (\lambda_p \cdot \mathcal{L}_{\text{physics}} + \lambda_r \cdot\mathcal{L}_{\text{reg}} + \lambda_{ort} \cdot \mathcal{L}_{\text{ort}}),
\end{equation}
where the physical loss has been presented in Equation \ref{eq6}, the regularization loss can be found in our previous work \citep{cheng2025meta}, $\lambda_p$, $\lambda_r$, and $\lambda_{ort}$ are the hyperparameters controlling the weight of the corresponding losses, and $\lambda_{scale}$ denotes a scaling factor that is employed to adjust the magnitude of the total loss value. We emphasize that we use this loss function to optimize network parameters in both meta-training and meta-testing stages. To make the meta-training process more stable, we set $\lambda_{scale}$ to 0.1. In the meta-testing stage, we inherit this setting. Meanwhile, to demonstrate the generalizablity of our Meta-LRPINN, we do not fine-tune the optimal configurations of $\lambda_p$, $\lambda_r$, and $\lambda_{ort}$. Instead, they are all set to 1.

By incorporating the orthogonality loss, the proposed loss function not only enforces the physical constraints of the wavefield and prevents overfitting, but also ensures the correctness of the SVD-based low-rank representation. This comprehensive loss design enhances the robustness and interpretability of the LRPINN while maintaining computational efficiency.
\section{\textbf{Numerical examples}}\label{examples}
In this section, we will validate the proposed Meta-LRPINN framework through a series of numerical examples. We begin by providing details of the meta-training stage, including the dataset construction, and the network and the training configurations. Then, we evaluate the performance of our method on two distinct models. First, we share our test on a layered velocity model to assess its ability to handle simple subsurface configurations and multi-frequency wavefields. Next, we apply the framework to the overthrust model, a more complex and realistic geological scenario, to demonstrate its scalability and adaptability. Finally, we evaluate the effectiveness of our rank-adaptive reduction strategy, highlighting its impact on computational efficiency and model performance.

\subsection{Meta-training procedure}\label{example1}
As stated earlier, the meta-training stage is designed to optimize the initialization parameters of the LRPINN and the FEH across various velocity models and a range of frequencies, enabling rapid adaptation to new velocity models and frequency configurations during the meta-testing phase. To achieve this, we construct a diverse set of tasks that encompass varying geological scenarios and frequency characteristics.

We generate a total of 40 velocity models with velocities ranging from 1.5 km/s to 5 km/s. The velocity models vary in size, with dimensions ranging from 2 km $\times$ 2 km (smallest) to 5 km × 20 km (largest). This range ensures the inclusion of both compact and large-scale geological settings. For each velocity model, we consider 14 discrete frequencies between 2 and 15 Hz, resulting in a total of 560 tasks. To generate training data, we randomly sample 40000 spatial points within the domain, where the samples include spatial coordinates $x$ and $z$, the corresponding source's location $x_s$, the velocity $v$, and the background velocity $v_0$ at the random drawn location. All the source corresponding the training data, as well as the following meta-testing data, are fixed in a depth of 0.025 km. 

The LRPINN, in addition to the input and output layers, consists of six hidden layers, where the weight matrices in each hidden layer are decomposed into low-rank representations. Specifically, the decomposed matrices $U_l$ and $V_l$ of each hidden layer are of size 320$\times$100, while the singular value matrix $\Theta_l$ is 100$\times$100. This corresponds to a fixed rank of 100 across all layers, balancing expressiveness and computational efficiency. The FEH network, responsible for predicting frequency-dependent singular values, consists of three fully connected layers with 80 neurons per layer. 

The meta-training process is conducted for a total of 50000 epochs using an AdamW optimizer \citep{loshchilov2017decoupled}. In each epoch, we sample five support-query task pairs along with their corresponding frequencies from the created 560 tasks (sets of velocity models and frequencies). The inner loop, responsible for task-specific adaptation, uses a learning rate of 2e-3 and performs a single gradient descent step per iteration. The outer loop updates the global initialization parameters with a learning rate of 1e-3, which decayed by a factor of 0.8 every 5000 epochs to ensure convergence. The meta-training phase is conducted on an NVIDIA A100 GPU [80 GB]. The entire training process takes approximately 34 hours. 

\subsection{A layered model}
\begin{figure}[htbp]
\centering
\includegraphics[width=0.5\textwidth]{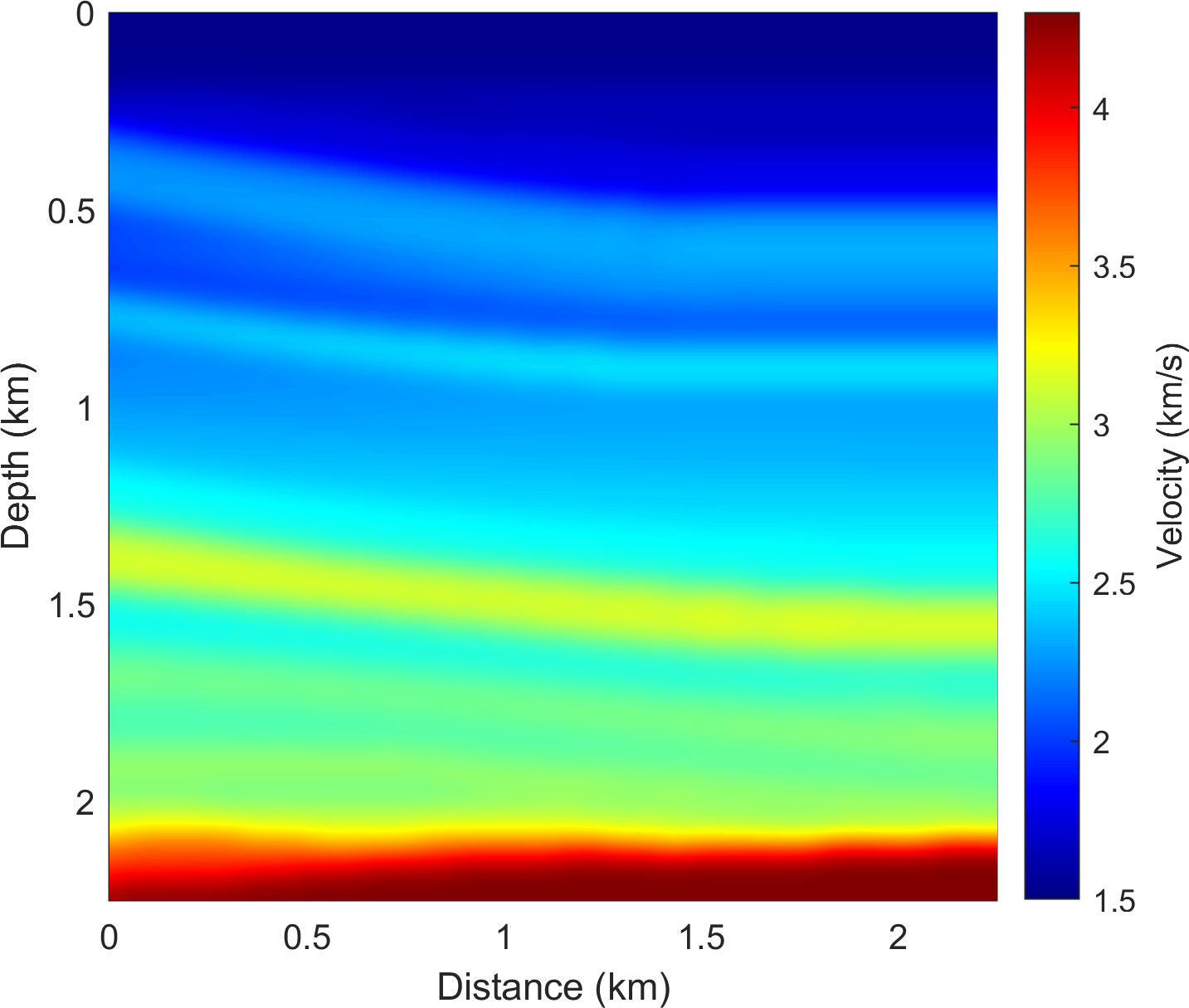}
\caption{The layered velocity model extracted from the Marmousi model.}
\label{fig2}
\end{figure}

To demonstrate the effectiveness of our proposed Meta-LRPINN framework, we first evaluate it on a layered velocity model extracted from the Marmousi model. Importantly, this model is not included in the meta-training stage, ensuring a fair assessment of the generalization capability of our method. Figure~\ref{fig2} shows the layered velocity model, which spans a domain of 2.25 km $\times$ 2.25 km. For this test, we leverage the initialization obtained from the meta-training stage to start the meta-testing process. The networks are trained for three frequencies, including 3 Hz, 6 Hz, and 12 Hz, with each frequency corresponding to a separate Meta-LRPINN network. 

To provide meaningful benchmarks, we compare our Meta-LRPINN against two baseline methods: Meta-PINN and vanilla PINN. The Meta-PINN framework also undergoes a meta-training process using the same dataset and configurations described in the previous subsection. For fairness, both vanilla PINN and Meta-PINN employ six hidden layers, identical to the configuration of Meta-LRPINN, while each hidden layer has 320 neurons. During meta-testing of Meta-LRPINN and Meta-PINN and, also, the training of vanilla PINN, we use the same setup, including 40000 randomly sampled spatial points, an initial learning rate of $1e-3$, and the AdamW optimizer. The learning rate is decayed by a factor of 0.5 at 2000th, 4000th, and 8000th epochs. This standard setup is chosen to ensure a fair comparison and to test the general applicability of our Meta-LRPINN framework. 

\begin{figure}[htbp]
\centering
\includegraphics[width=0.95\textwidth]{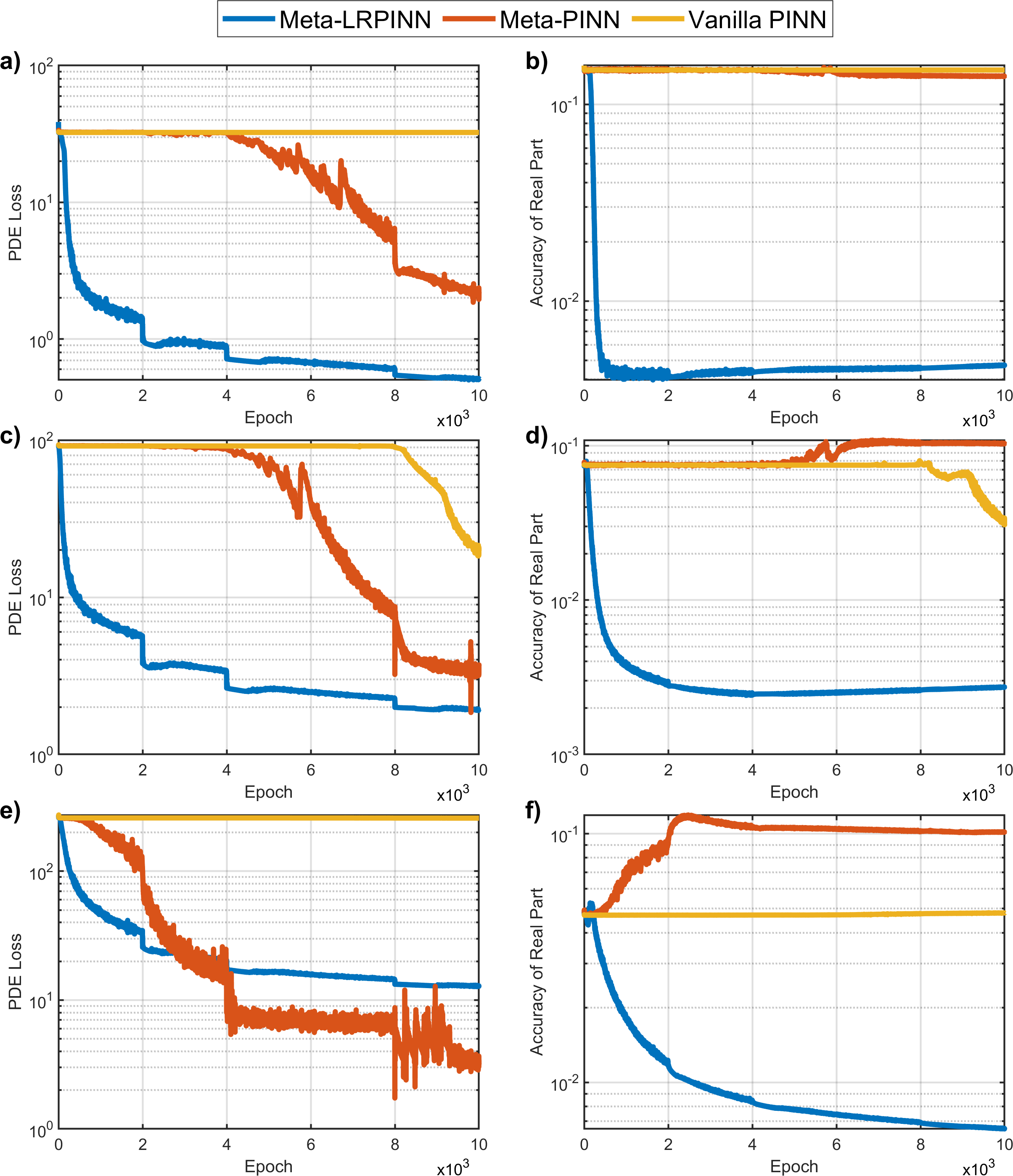}
\caption{Comparison of physical loss and accuracy curves between our Meta-LRPINN (blue), Meta-PINN (orange), and vanilla PINN (yellow) on the layered velocity model. The rows correspond to different frequencies (top: 3 Hz, middle: 6 Hz, bottom: 12 Hz). For each frequency, the left column shows the physical loss curves, and the right column presents the accuracy curves (measured as MSE against the numerical reference).}
\label{fig3}
\end{figure}

Figures~\ref{fig3} presents comparisons of physical loss and accuracy curves between our Meta-LRPINN and two benchmarks. Here, the accuracy is calculated using the mean squared error (MSE) between the network-predicted wavefields and the numerical references, where the numerical references are computed using a finite-difference method. In this figure, the physical loss and accuracy curves are shown for 3 Hz (top row), 6 Hz (middle row), and 12 Hz (bottom row), with physical loss on the left and accuracy on the right for each frequency. We can see that our Meta-LRPINN significantly outperforms both Meta-PINN and vanilla PINN in terms of physical loss for 3 Hz and 6 Hz. At 12 Hz, although Meta-LRPINN initially exhibits better performance than Meta-PINN, the loss reduction for Meta-PINN accelerates during later epochs. However, the accuracy curves reveal that our Meta-LRPINN consistently outperforms both benchmarks across all frequencies. Notably, the accuracy of Meta-PINN decreases as training progresses for 12 Hz, suggesting that it converges to a trivial solution or a local minimum, despite the observed loss reduction. 

\begin{figure}[htbp]
\centering
\includegraphics[width=1\textwidth]{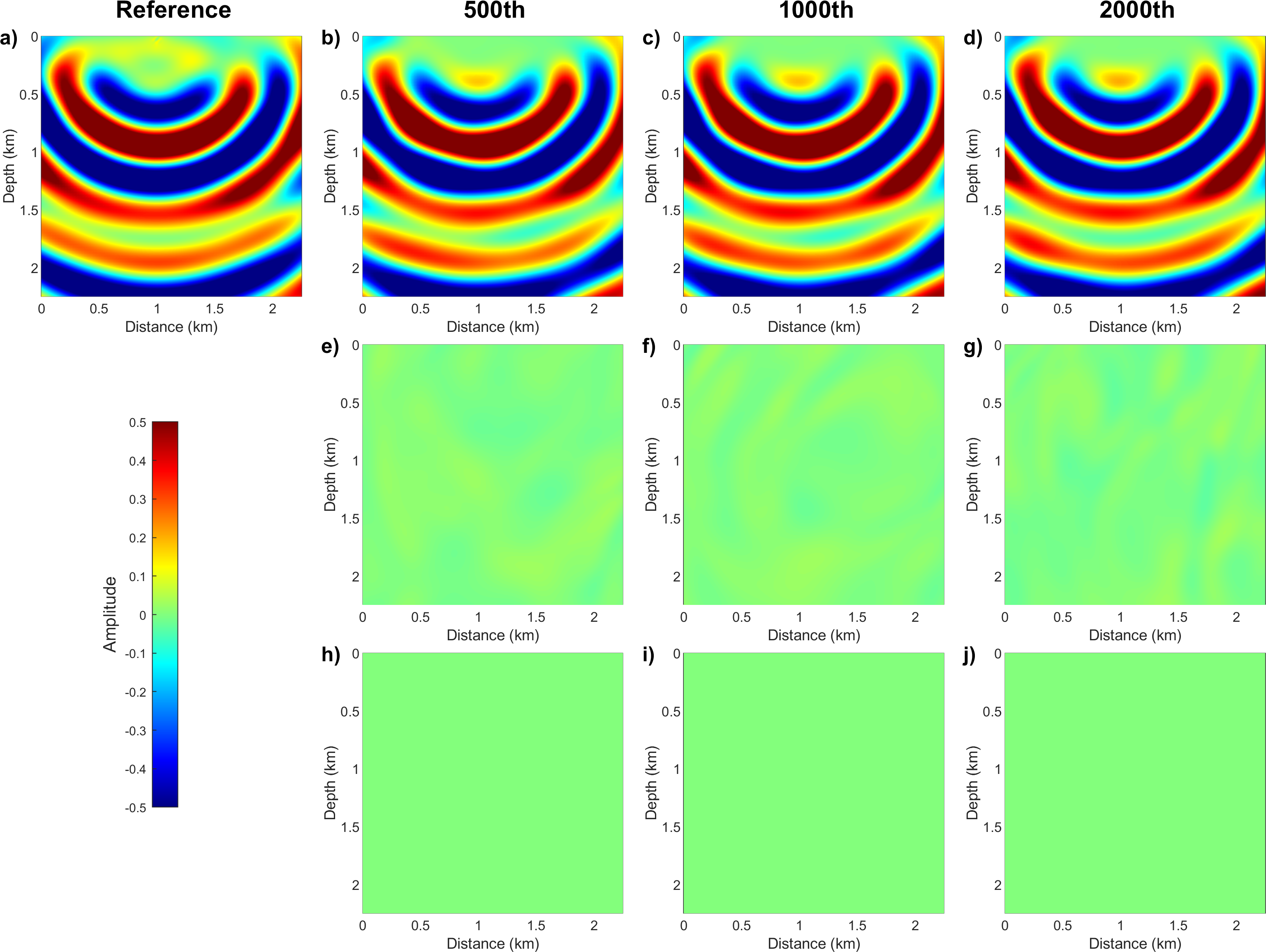}
\caption{Comparison of the real part of the scattered wavefield solutions at 3 Hz for the layered velocity model. (a) Numerical reference solution. Subsequent rows represent wavefields predicted by Meta-LRPINN, Meta-PINN, and vanilla PINN, respectively. Columns correspond to different training epochs, where the specific epoch numbers are indicated in the top.}
\label{fig4}
\end{figure}

\begin{figure}[htbp]
\centering
\includegraphics[width=1\textwidth]{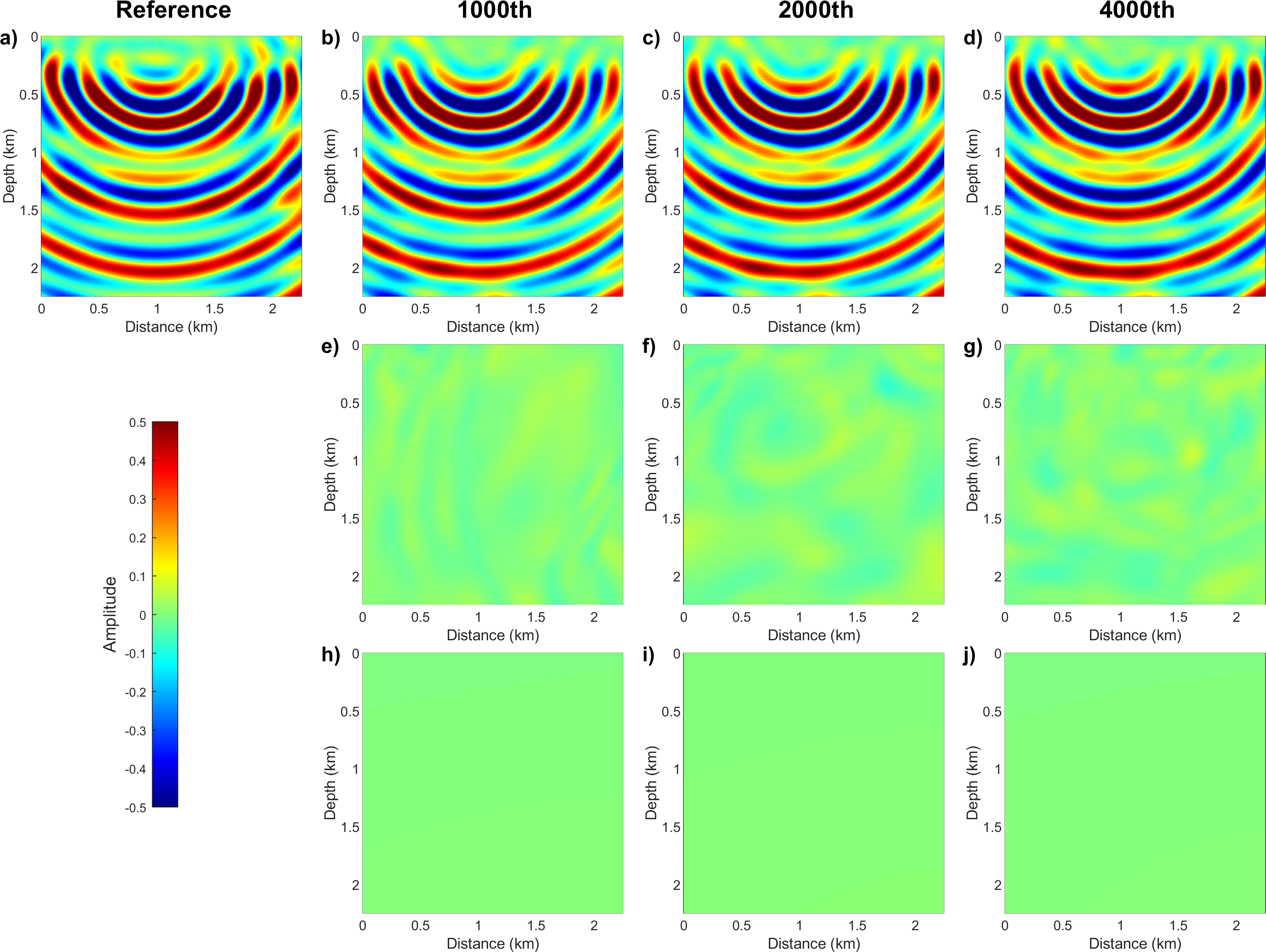}
\caption{Similar with Figure \ref{fig4}, but for the scattered wavefield solutions of 6 Hz.}
\label{fig5}
\end{figure}

\begin{figure}[htbp]
\centering
\includegraphics[width=1\textwidth]{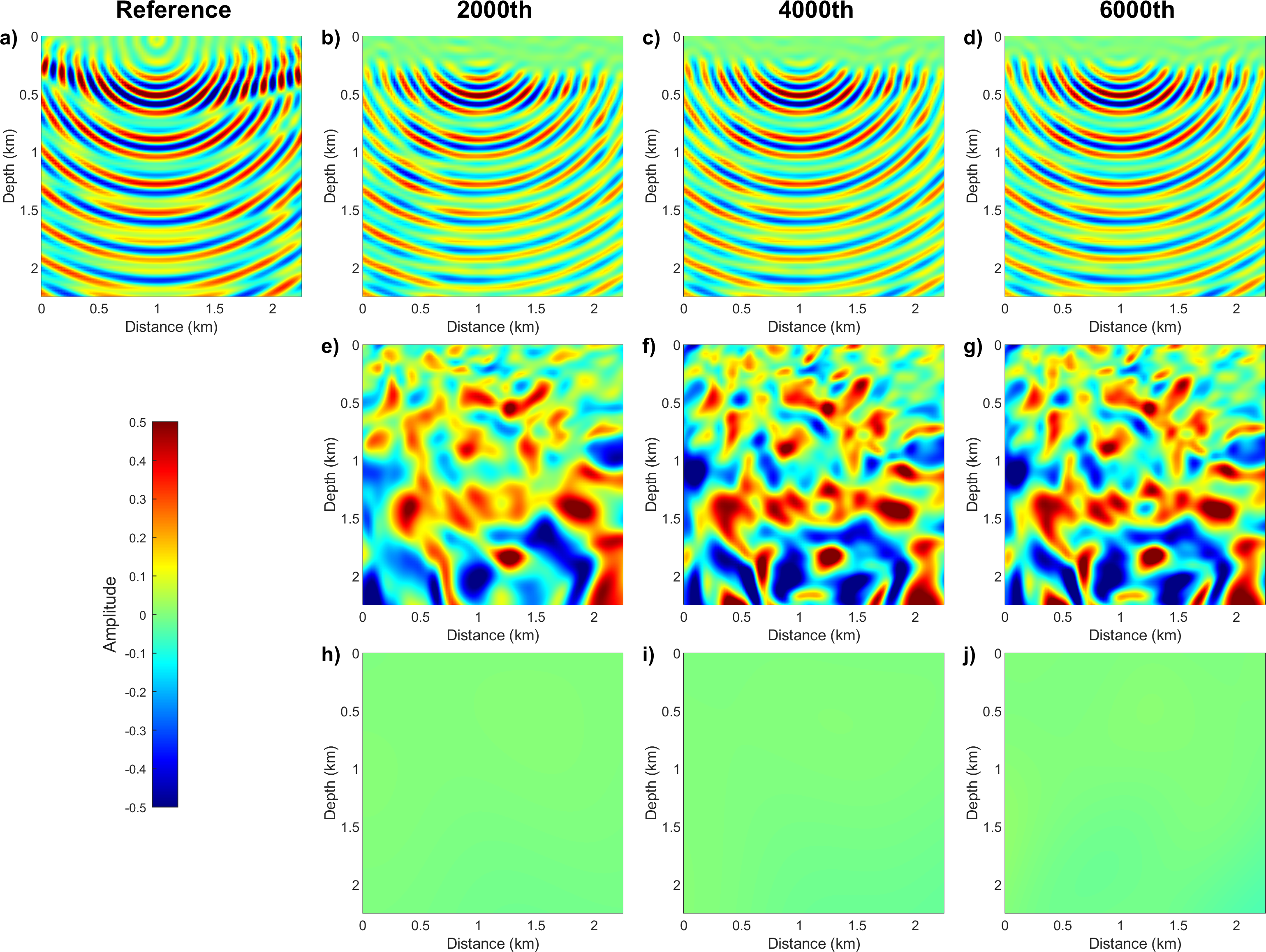}
\caption{Similar with Figure \ref{fig4}, but for the scattered wavefield solutions of 12 Hz.}
\label{fig6}
\end{figure}

After training, we evaluate the NN multi-frequencies wavefield representations for the given layered velocity model.  Figures~\ref{fig4}, \ref{fig5}, and \ref{fig6} provide visual comparisons of the real part of the predicted scattered wavefields at 3 Hz, 6 Hz, and 12 Hz, respectively. For brevity, the imaginary parts are omitted here, as well as in the following tests, to avoid redundant presentation. In each figure, panel (a) shows the numerical reference wavefield, while subsequent rows correspond to Meta-LRPINN, Meta-PINN, and vanilla PINN predictions. Each column represents different training epochs, with specific epoch numbers indicated in the top. These figures clearly demonstrate that Meta-LRPINN provides reliable wavefield representations after very limited training epochs, even for higher frequencies. In contrast, neither Meta-PINN nor vanilla PINN produces acceptable prediction results under similar conditions. For instance, Meta-LRPINN achieves near-perfect wavefield representation by the 500th epoch for 3 Hz, while Meta-PINN and vanilla PINN fail to converge to accurate solutions within the tested training epochs.

\subsection{Overthrust model}
\begin{figure}[htbp]
\centering
\includegraphics[width=0.8\textwidth]{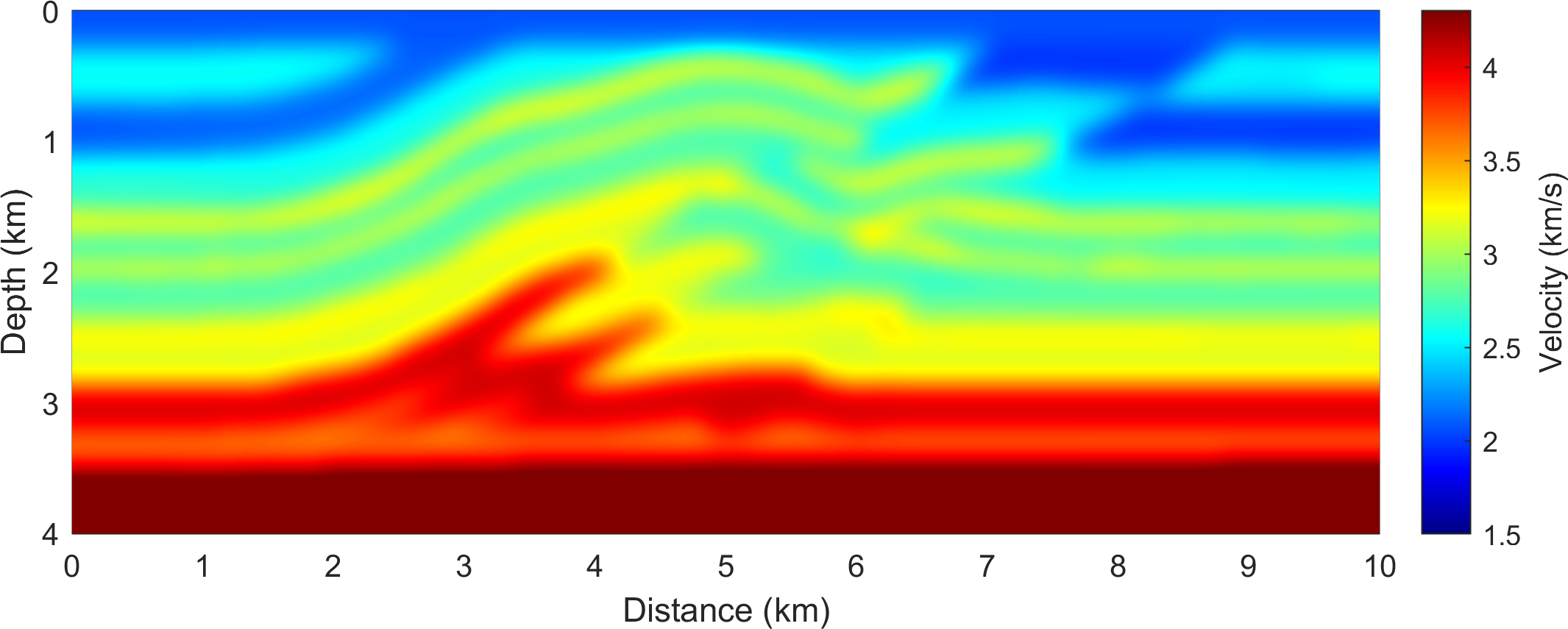}
\caption{Overthrust velocity model.}
\label{fig7}
\end{figure}

We further evaluate the performance of Meta-LRPINN on the overthrust model, a more complex velocity with uneven dimensions of 4 km $\times$ 10 km, which is usually a challenge for PINNs. A slightly smoothed version of the velocity model is shown in Figure \ref{fig7}. Similar to the layered model experiment, we, here, also consider three frequencies, 3 Hz, 6 Hz, and 12 Hz, and train separate (already meta-trained) Meta-LRPINN networks for each frequency. We compare the results against two benchmarks: Meta-PINN and vanilla PINN. The training configurations for all three methods remain consistent with those used in the layered model experiments. 

\begin{figure}[htbp]
\centering
\includegraphics[width=0.95\textwidth]{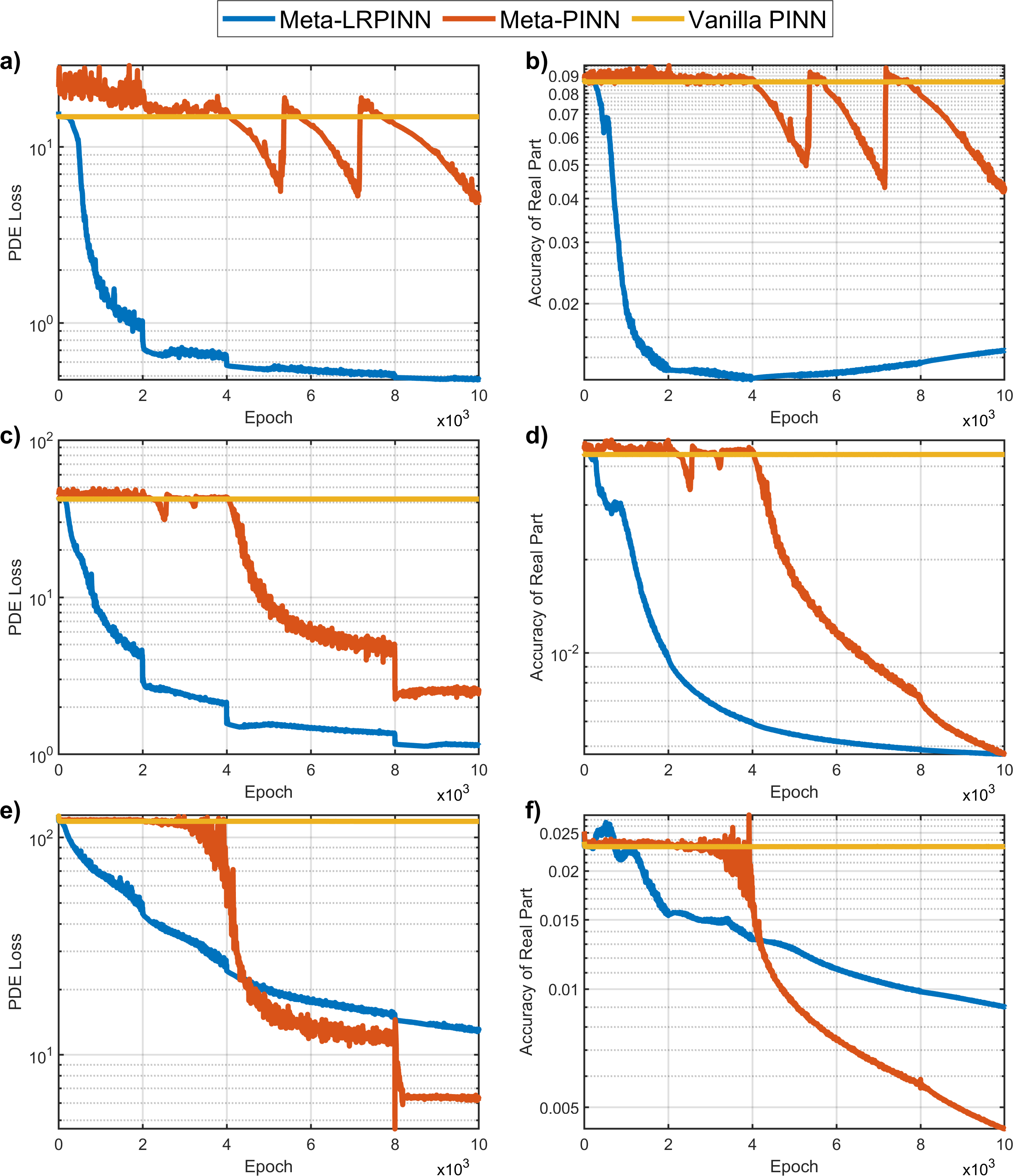}
\caption{Comparison of physical loss and accuracy curves between our Meta-LRPINN (blue), Meta-PINN (orange), and vanilla PINN (yellow) on the overthrust model. The rows correspond to different frequencies (top: 3 Hz, middle: 6 Hz, bottom: 12 Hz). For each frequency, the left column shows the physical loss curves, and the right column presents the accuracy curves (measured as MSE against the numerical reference).}
\label{fig8}
\end{figure}

Figure \ref{fig8} presents the PDE loss and accuracy curves for the three frequencies, with rows corresponding to 3 Hz, 6 Hz, and 12 Hz, and columns showing the PDE loss (left) and the accuracy (right). The results demonstrate that Meta-LRPINN significantly outperforms both benchmarks at 3 Hz and 6 Hz in terms of convergence speed and accuracy. For 12 Hz, Meta-LRPINN initially shows better performance than Meta-PINN; however, as training progresses, Meta-PINN exhibits faster reductions in loss and accuracy improvements. This suggests that for representing high-frequency wavefields in large models, Meta-LRPINN may require a higher rank to further enhance performance. 

\begin{figure}[htbp]
\centering
\includegraphics[width=1\textwidth]{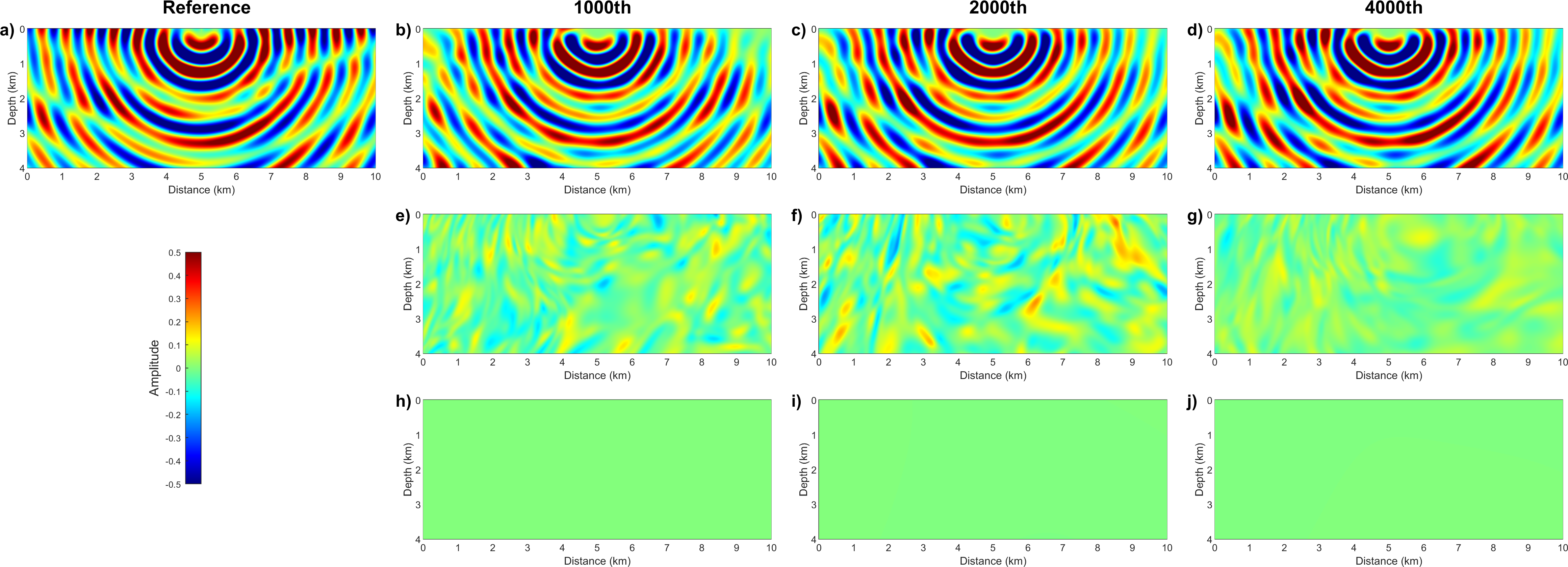}
\caption{Comparison of the real part of the scattered wavefield solutions at 3 Hz for the overthrust model. (a) Numerical reference solution. Subsequent rows represent wavefields predicted by Meta-LRPINN, Meta-PINN, and vanilla PINN, respectively. Columns correspond to different training epochs, where the specific epoch numbers are indicated in the top.}
\label{fig9}
\end{figure}

\begin{figure}[htbp]
\centering
\includegraphics[width=1\textwidth]{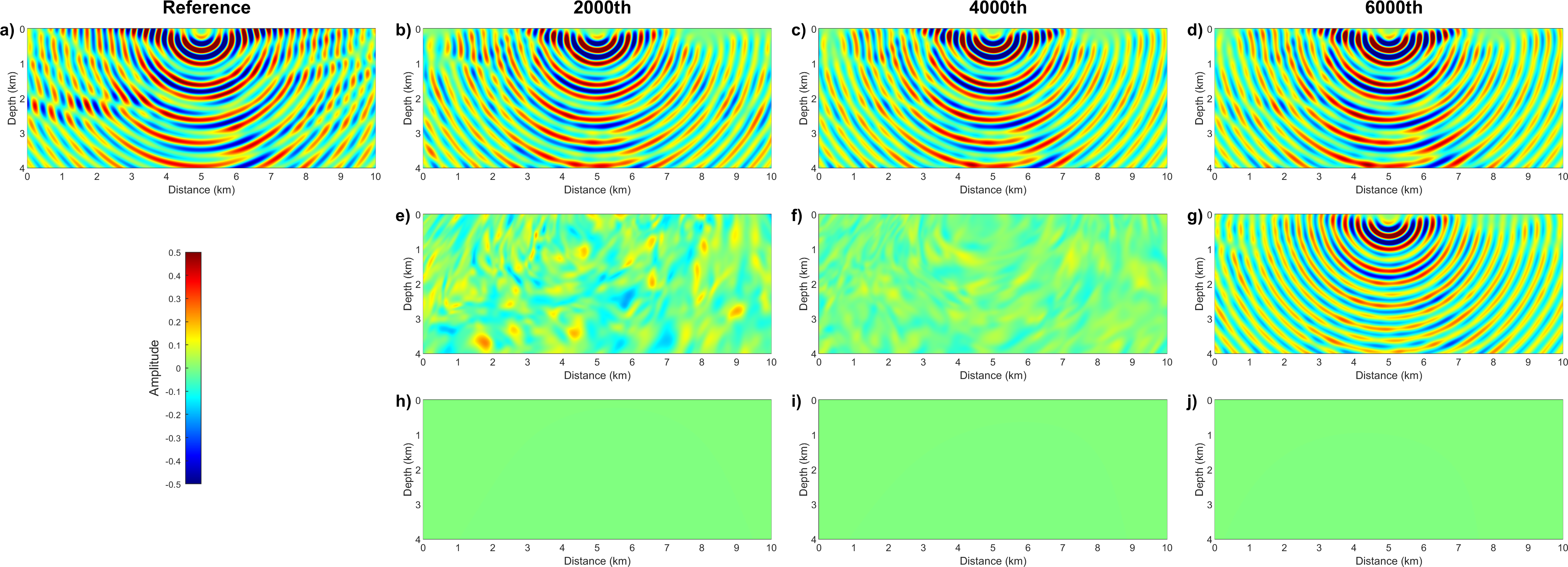}
\caption{Similar with Figure \ref{fig9}, but for the scattered wavefield solutions of 6 Hz.}
\label{fig10}
\end{figure}

\begin{figure}[htbp]
\centering
\includegraphics[width=1\textwidth]{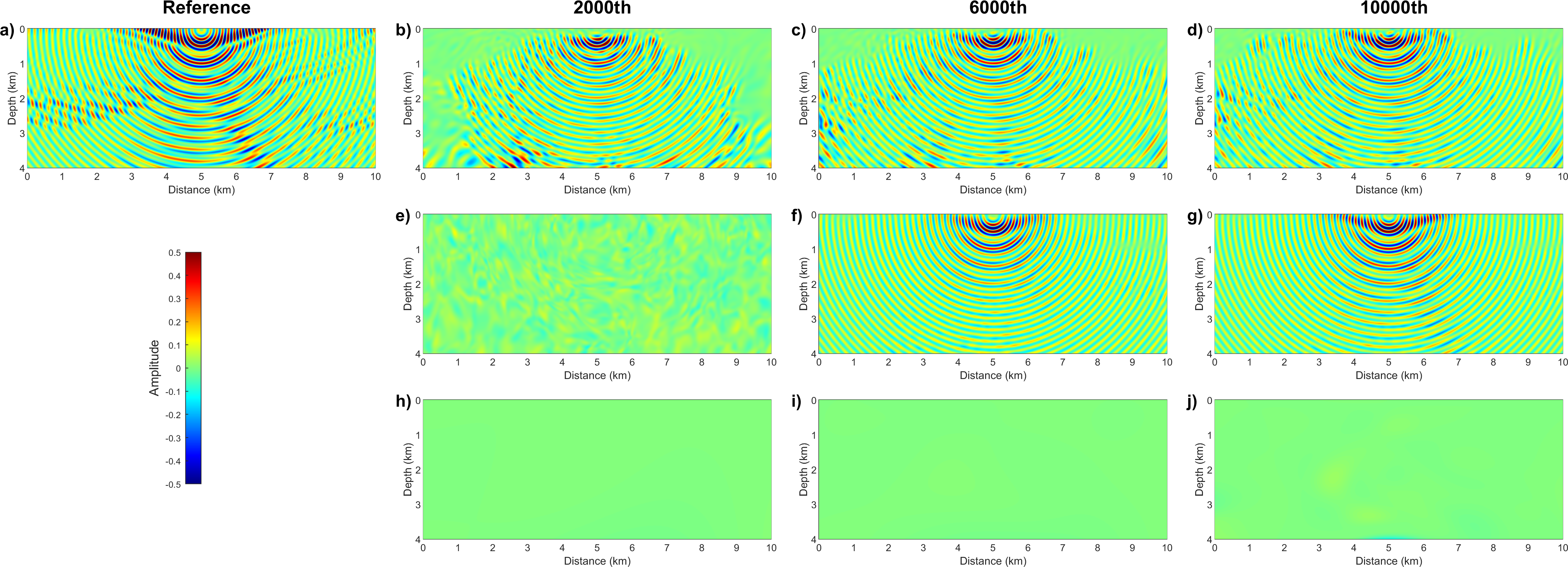}
\caption{Similar with Figure \ref{fig9}, but for the scattered wavefield solutions of 12 Hz.}
\label{fig11}
\end{figure}

Figures \ref{fig9}, \ref{fig10}, and \ref{fig11} compare the real part of the scattered wavefield solutions for 3 Hz, 6 Hz, and 12 Hz, respectively. Each figure follows the layout used in the layered model experiment: panel (a) displays the numerical reference, and subsequent rows correspond to Meta-LRPINN, Meta-PINN, and vanilla PINN. Columns depict the wavefield solutions at different training epochs, as indicated by the above label. 

From Figure \ref{fig9}, for the 3 Hz wavefield, Meta-LRPINN delivers a wavefield representation closely matching the reference as early as the 1000th epoch. In contrast, Meta-PINN and vanilla PINN struggles to converge, and even at the 4000th epoch, they fail to provide acceptable solutions. In Figure  \ref{fig10}, which shows the 6 Hz wavefield, Meta-LRPINN provides an acceptable wavefield representation by the 2000th epoch and continues to refine it with further training. Meta-PINN requires up to 6000 epochs to achieve a wavefield solution, but its details remain inferior to the solution provided by Meta-LRPINN at the 2000th epoch. Vanilla PINN again fails to produce any meaningful results. For the high-frequency 12 Hz wavefield shown in Figure \ref{fig11}, Meta-LRPINN achieves a moderately reasonable wavefield representation by the 2000th epoch. While Meta-PINN starts to show wavefield features only by the 6000th epoch, its representation lags behind the solution provided by Meta-LRPINN, especially for some details. Vanilla PINN remains incapable of modeling the high-frequency wavefield. However, the Meta-LRPINN overall could not capture the details in the reference wavefield solution. This is mainly because this wavefield is complex and it spans an uneven domain. Note in Figure \ref{fig10}, how well the prediction is for the parts deeper directly below the source, compared to the sides. Some of this limitation could be related to the rank of the weight SVD representation we chose, which will be discussed next.

\subsection{Rank reduction}
Finally, we evaluate the proposed rank-adaptive reduction strategy. We, here, use the layered velocity model as an example. As described in the Meta-training procedure, the rank for our Meta-LRPINN during meta-training is set to 100. While the results in the layered model test directly apply this configuration without rank reduction, here we explore the effect of reducing the rank during the meta-testing stage. We consider three levels of rank reduction relative to the meta-trained rank of 100: 50\% reduction (rank = 50), 75\% reduction (rank = 25), and 90\% reduction (rank = 10). The objective is to assess the impact of reduced ranks on the physical loss, accuracy, and wavefield representation. We also consider the wavefield representation at 3 Hz, 6 Hz, and 12 Hz, like we did in the layered model test. 

\begin{figure}[htbp]
\centering
\includegraphics[width=0.95\textwidth]{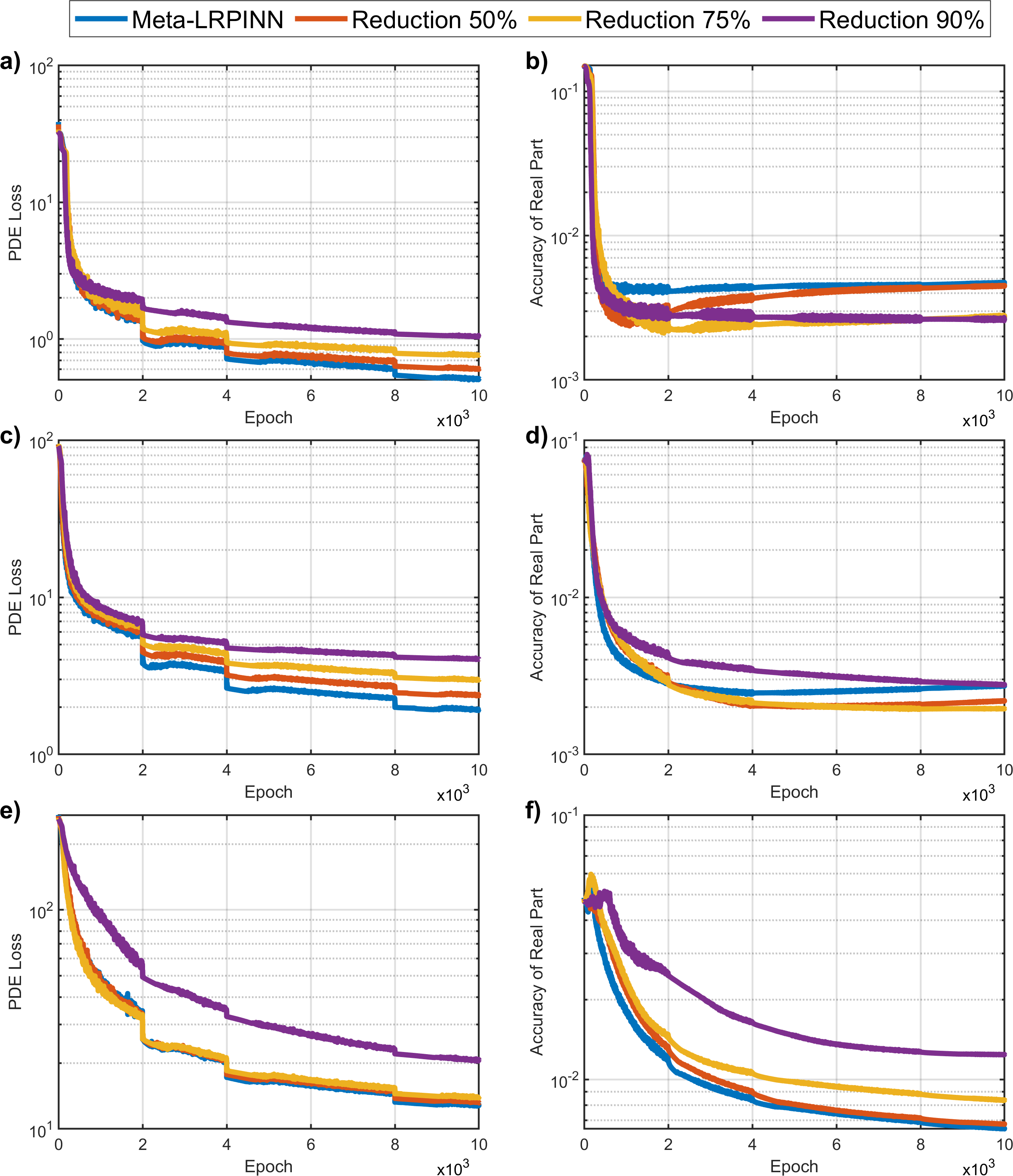}
\caption{Comparison of physical loss and accuracy curves between four rank configurations on the layered velocity model: rank = 100 (blue), 50\% reduction (orange), 75\% reduction (yellow), and 90\% reduction (purple). The rows correspond to different frequencies (top: 3 Hz, middle: 6 Hz, bottom: 12 Hz). For each frequency, the left column shows the physical loss curves, and the right column presents the accuracy curves (measured as MSE against the numerical reference).}
\label{fig12}
\end{figure}

Figure~\ref{fig12} compares the physical loss (left column) and accuracy (right column) curves for the four rank configurations, where each column from top to bottom corresponds to 3, 6, and 12Hz, respectively. From the physical loss curves, we can observe that as the rank decreases, the physical loss increases across all frequencies, with the most significant degradation occurring at the 90\% rank reduction level. The original rank-100 configuration consistently achieves the lowest physical loss, followed by the 50\% and 75\% reductions. Interestingly, at 12 Hz, the physical loss curves for the 50\% and 75\% reductions exhibit convergence speeds and loss values that are very close to the original rank-100 configuration. This suggests that moderate rank reductions retain sufficient capacity to minimize the physical loss even for high-frequency components, although further reductions (e.g., 90\%) introduce noticeable errors. 

The accuracy curves reveal interesting insights. We can see that, at 3 Hz, reducing the rank improves accuracy compared to the original rank-100 configuration. The 75\% reduction (rank = 25) achieves the highest accuracy, surpassing all other configurations, while the 90\% reduction (rank = 10) also performs well, with only a slight drop in accuracy compared to the 75\% reduction. This suggests that adaptively reducing the rank effectively regularizes the model, enhancing generalization for low-frequency wavefields. At 6 Hz, the accuracy curves reveal a two-stage trend: during the initial epochs, the rank-100 configuration exhibits much fast accuracy improvement. However, as training progresses, the 50\% and 75\% reductions surpass the rank-100 configuration. This further indicates that moderate rank reductions retain sufficient representational capacity while reducing overparameterization. However, at 12 Hz, the rank-100 configuration provides the best accuracy, with a clear decline in performance as the rank reduction increases. This is expected, as high-frequency wavefields contain finer details that require more singular values to accurately capture. Nonetheless, the 50\% reduction retains reasonable accuracy at 12 Hz, demonstrating that moderate reductions are still effective for certain high-frequency scenarios. 

\begin{figure}[htbp]
\centering
\includegraphics[width=1\textwidth]{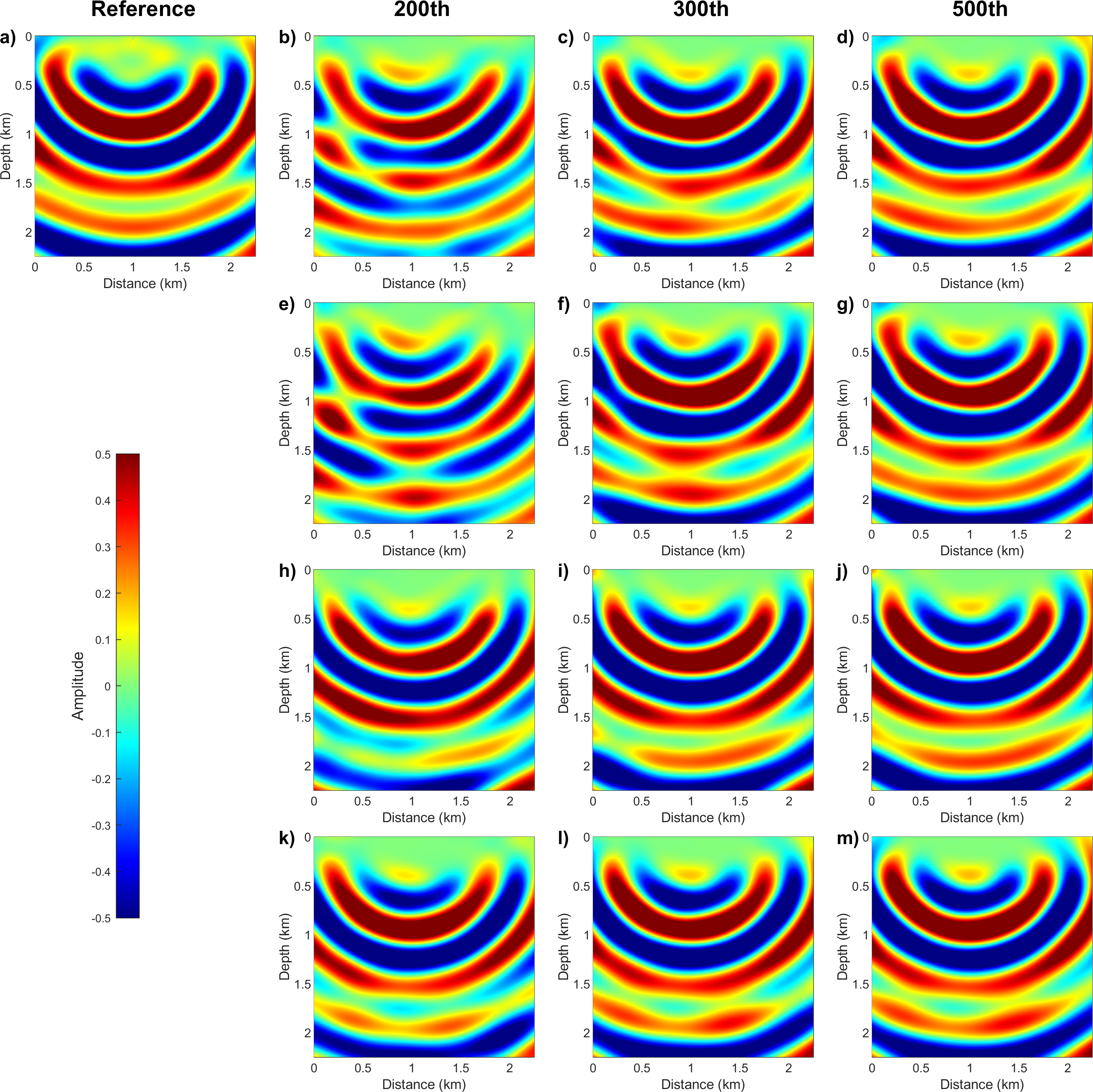}
\caption{Comparison of the real part of the scattered wavefield solutions at 3 Hz for the layered velocity model (Figure \ref{fig2}), where we use four rank configurations. (a) Numerical reference solution. Subsequent rows represent wavefields predicted by: rank = 100 (blue), 50\% reduction (orange), 75\% reduction (yellow), and 90\% reduction (purple). Columns correspond to different training epochs, where the specific epoch numbers are indicated on the top.}
\label{fig13}
\end{figure}

\begin{figure}[htbp]
\centering
\includegraphics[width=1\textwidth]{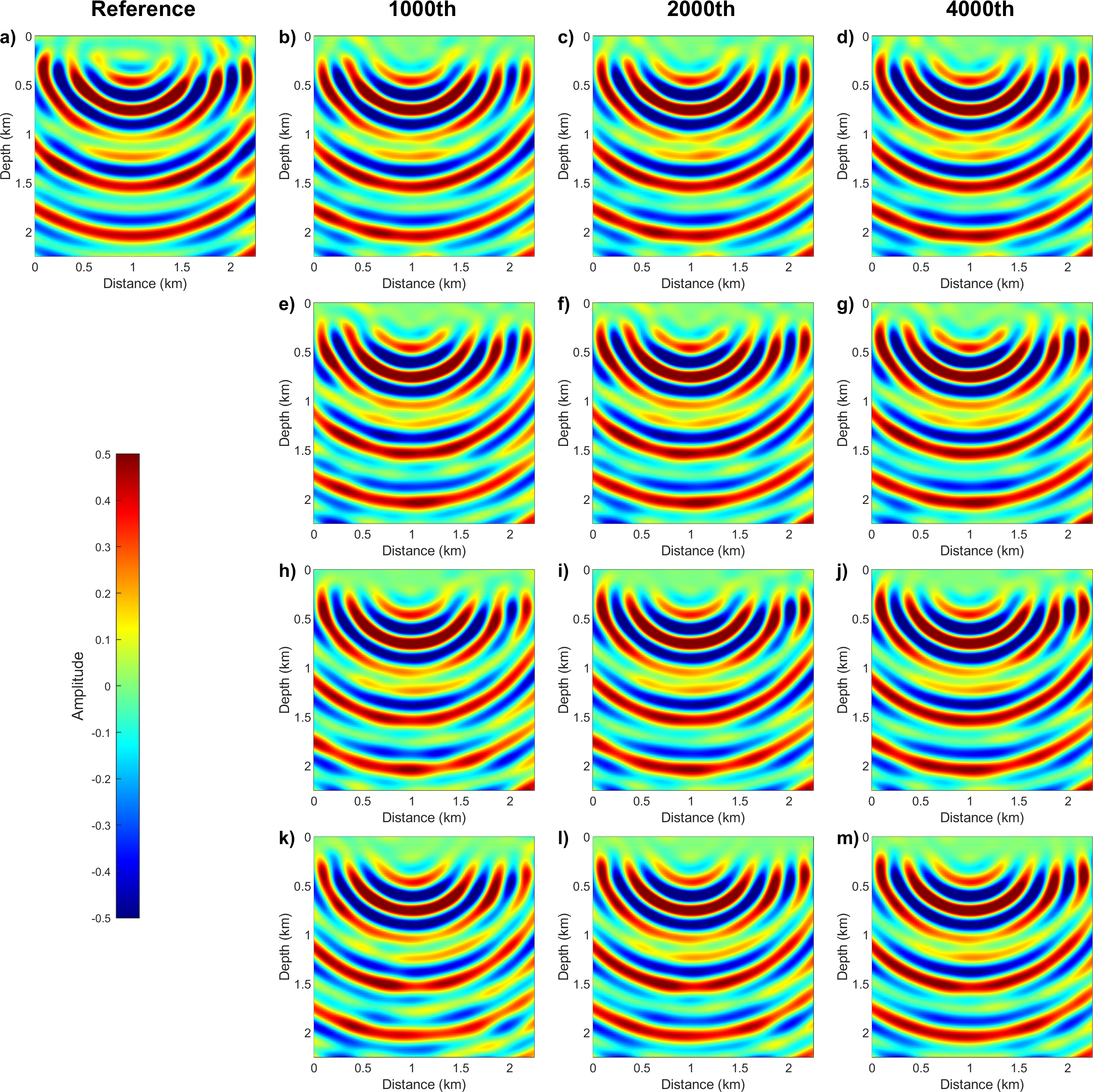}
\caption{Similar with Figure \ref{fig13}, but for the scattered wavefield solutions of 6 Hz.}
\label{fig14}
\end{figure}

\begin{figure}[htbp]
\centering
\includegraphics[width=1\textwidth]{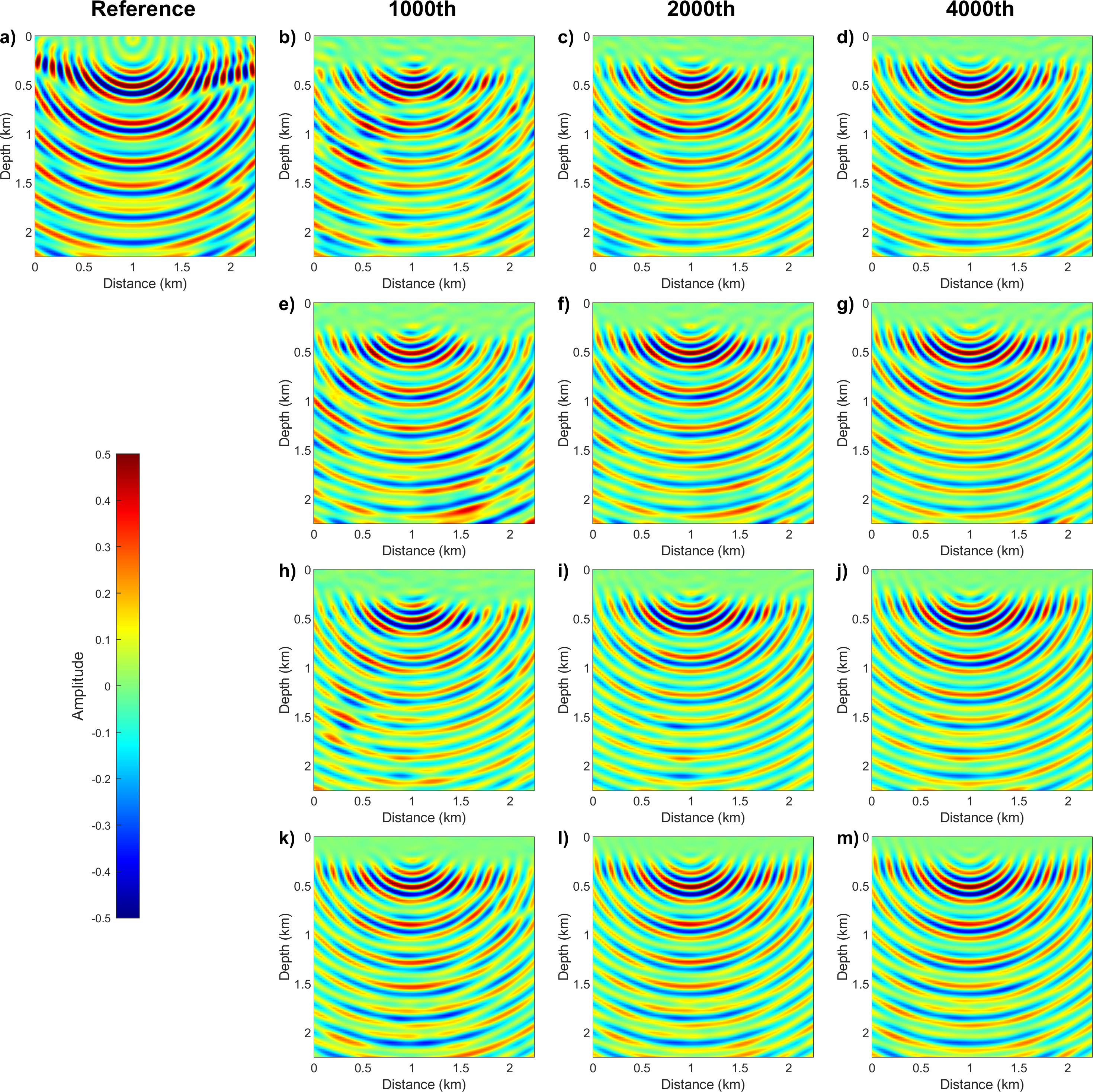}
\caption{Similar with Figure \ref{fig13}, but for the scattered wavefield solutions of 12 Hz.}
\label{fig15}
\end{figure}

Figures~\ref{fig13}, \ref{fig14}, and \ref{fig15} further illustrate the impact of rank reduction on the real part of the scattered wavefields at 3 Hz, 6 Hz, and 12 Hz, respectively. Each figure consists of: Panel (a) corresponds to numerical reference wavefield. Panels (b–m): Predicted wavefields at different epochs (indicated above each column) under the four rank configurations, where each column from top to bottom corresponds to rank = 100, rank = 50 (50\% reduction), rank = 25 (75\% reduction), and rank = 10 (90\% reduction), respectively. 

For the 3 Hz low-frequency wavefield representation, rank reductions of 75\% and 90\% demonstrate the ability to quickly provide more accurate wavefield approximations. By the 200th epoch, the wavefields predicted by the 75\% and 90\% rank reductions are significantly closer to the reference wavefield compared to the original rank-100 and 50\% rank reduction models, with the 90\% reduction producing wavefields that closely match the reference in both phase and amplitude. In contrast, the rank-100 and 50\% reduction models still exhibit noticeable differences from the reference wavefield at this early stage. As training progresses to the 500th epoch, all configurations, including the rank-100 and three rank-reduction models, converge to produce wavefields that visually resemble the reference wavefield, making it difficult to discern any significant differences among them. This suggests that for low frequencies, rank reductions can achieve faster convergence with fewer parameters while still maintaining accuracy. 

At 6 Hz, the wavefield representation exhibits different characteristics. During the initial epochs, such as at the 1000th epoch, the original rank-100 configuration provides wavefields that are closer to the reference, particularly in the deeper regions of the wavefield where details are more prominent. Rank reductions, especially the 90\% reduction, show some limitations in representing these deep wavefield details. These limitations are particularly evident in regions with lower amplitudes, where smaller singular values are responsible for capturing fine-grained information. Removing a significant portion of these singular values in aggressive rank reductions (e.g., 90\%) results in a less detailed representation of the deeper wavefields. However, as training progresses, the rank-100 model and the 50\% and 75\% reductions converge to provide comparable wavefield representations, with all three showing better detail retention in the deeper regions compared to the 90\% reduction model. This highlights that moderate rank reductions are effective for mid-frequency wavefields, but aggressive reductions may compromise fine-detail representation. 

At 12 Hz, the high-frequency wavefield representation highlights the necessity of maintaining higher ranks. The original rank-100 model provides the best match to the reference wavefield throughout the training process, particularly in capturing the intricate details of the high-frequency components. As the rank is reduced, the ability to accurately represent these details decreases, with the 75\% and 90\% reductions exhibiting noticeable wavefield deformation. However, it is important to note that even with significant rank reductions (e.g., 90\%), the degradation in wavefield representation remains relatively mild given the corresponding substantial reduction in model parameters. This suggests that while retaining a higher rank is critical for accurately representing high-frequency wavefields, moderate reductions can still provide acceptable results for applications where computational efficiency is a priority. 

\section{\textbf{Discussion}}\label{discussion}
In this section, we will analyze and evaluate some key components of the proposed Meta-LRPINN framework to provide a deeper understanding of its performance and contributions. First, we discuss the role of meta-learning in providing an effective initialization for LRPINN. Second, we explore the impact of rank size on the performance of meta-LRPINN, highlighting how different rank configurations influence the accuracy, convergence, and computational efficiency. Third, we analyze the decision to prune the frequency embedding hypernetwork (FEH) during the meta-testing stage. Lastly, we examine the generalization capability of Meta-LRPINN to out-of-distribution frequencies, given that the meta-trained initialization is learned from a specific range of velocity and frequency distributions. 

\subsection{The contributions of meta-learning}
\begin{figure}[htbp]
\centering
\includegraphics[width=0.95\textwidth]{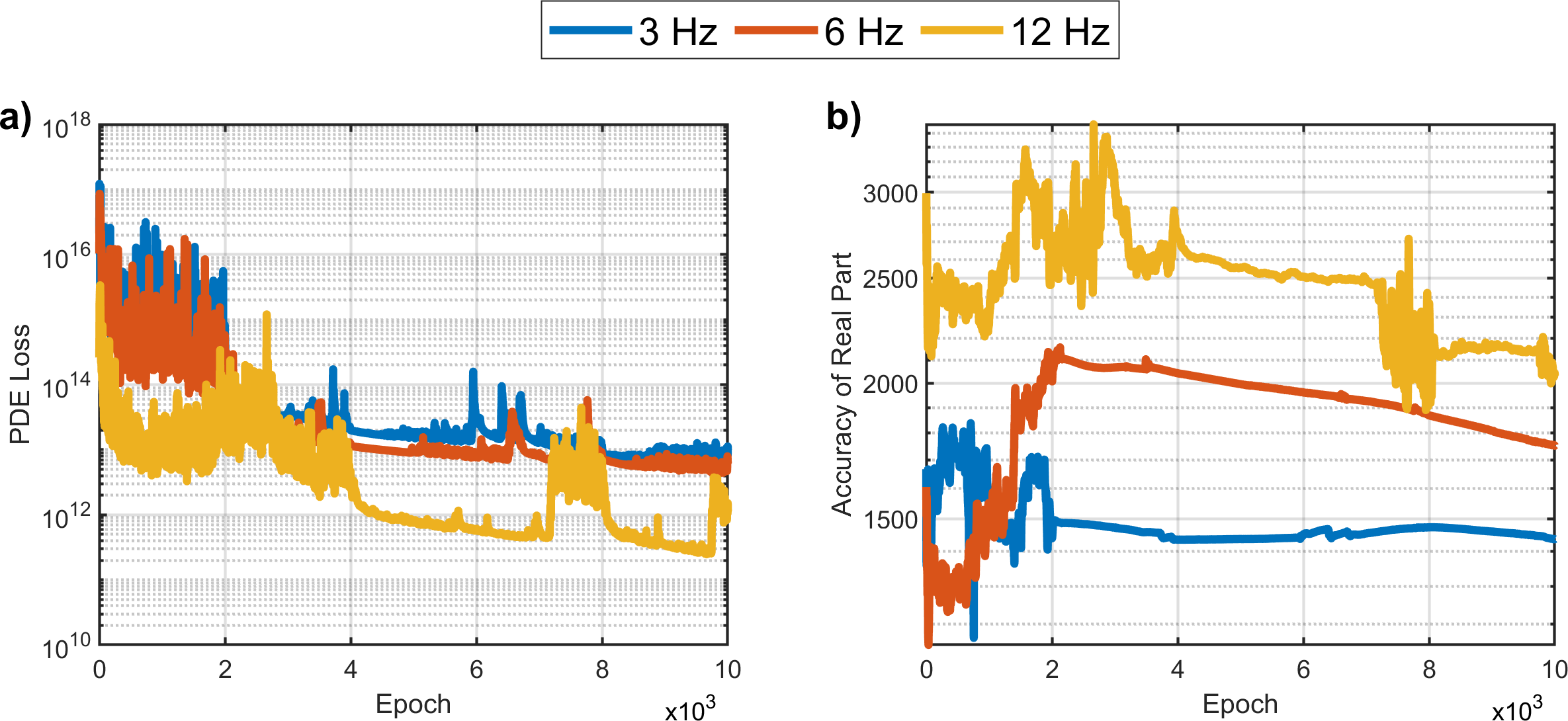}
\caption{The physical loss and accuracy curves of LRPINN initialized with random parameters for three different frequencies: 3 Hz (blue), 6 Hz (orange), and 12 Hz (yellow). (a) The physical loss curves. (b) The accuracy curves.}
\label{fig16}
\end{figure}

To evaluate the contribution of meta-learning to our Meta-LRPINN framework, we test the performance of randomly initialized LRPINN. Using the layered velocity model as an example, we consider three different frequencies, 3 Hz, 6 Hz, and 12 Hz, and train LRPINN directly from random initialization. The results are shown in Figure \ref{fig16}, where the PDE loss and accuracy curves for the three frequencies are presented, with panel (a) depicting the physical loss curves and panel (b) showing the accuracy curves. 

The results highlight the difficulty of optimizing LRPINN without meta-learned initialization. The PDE loss for the randomly initialized LRPINN across all three frequencies remains exceedingly high, ranging between $10^{12}$ and $10^{16}$, indicating the network struggles to minimize the physical loss during training. Similarly, the accuracy curves reveal that the error relative to the numerical reference is large, with values between 1500 and 3000 across all three frequencies. This is in stark contrast to the results shown in Figure \ref{fig3} for Meta-LRPINN with meta-learned initialization, where the loss and accuracy curves quickly converge to low values. These results demonstrate the critical role of meta-learning in providing robust initialization for LRPINN. By leveraging meta-learning, the optimization process is significantly stabilized, and the training efficiency is greatly enhanced, particularly for challenging tasks such as representing multi-frequency wavefields in variable velocity models.

\subsection{The effect of rank size}
\begin{figure}[htbp]
\centering
\includegraphics[width=0.95\textwidth]{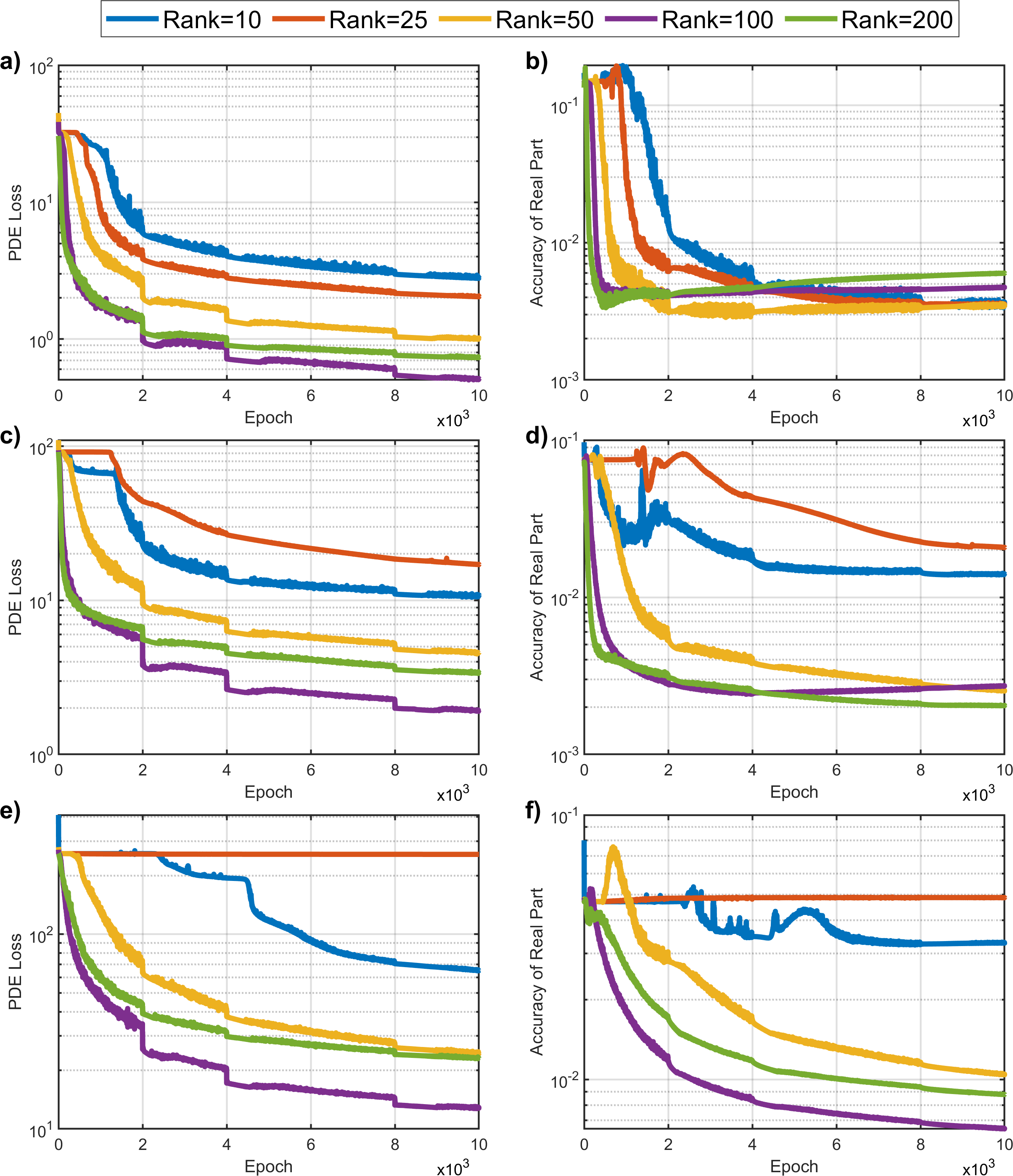}
\caption{Comparison of physical loss and accuracy curves between different rank sizes on layered model: rank = 10 (blue), 25 (orange), 50 (yellow), 100 (purple), and 200 (green). The rows correspond to different frequencies (top: 3 Hz, middle: 6 Hz, bottom: 12 Hz). For each frequency, the left column shows the physical loss curves, and the right column presents the accuracy curves (measured as MSE against the numerical reference).}
\label{fig17}
\end{figure}

To explore the influence of rank on the performance of Meta-LRPINN, we conduct an experiment where the rank is fixed during both the meta-training and meta-testing phases, as opposed to using the rank-adaptive reduction strategy in previous numerical examples. Specifically, we train Meta-LRPINNs with different ranks, including 10, 25, 50, 100, and 200, using the same meta-training configuration, with the only difference being the rank size. These models are then evaluated on the layered velocity model during the meta-testing phase for three frequencies: 3 Hz, 6 Hz, and 12 Hz. For each frequency, a separate network is trained. The results are presented in Figure~\ref{fig17}, which shows the physical loss and accuracy curves for each rank size across the three frequencies. Each column corresponds to a specific frequency, with rows displaying the physical loss (left) and accuracy (right) curves. 

From the physical loss curves, we observe no universal trend across all ranks and frequencies. However, certain patterns emerge. The rank-100 model consistently achieves the fastest convergence and the lowest physical loss across all frequencies. This is followed by the rank-200 and rank-50 models, which also exhibit competitive loss values. The performance of the rank-10 and rank-25 models varies: at 3 Hz, the rank-10 model performs worse than rank-25, while at 6 Hz and 12 Hz, rank-10 outperforms rank-25 in terms of both convergence speed and final loss values. This indicates that at higher frequencies, a smaller rank (e.g., rank-10) introduces a form of regularization that aids in convergence, although this effect diminishes for very high ranks such as 100 and 200, where the increased parameterization compensates for the lack of regularization. 

The accuracy curves provide additional insights into the impact of rank size. At 3 Hz, we can observe a clear trend where the convergence speed improve as the rank size increases. However, the final accuracy of all rank configurations holds almost the same. At 6 Hz, this trend is largely preserved, except that the rank-10 model demonstrates slightly faster accuracy convergence than rank-25. At 12 Hz, the rank-100 model achieves the best accuracy and convergence speed, followed by rank-200 and rank-50. The rank-10 and rank-25 models perform relatively poor, but rank-10 exhibits slightly stronger accuracy improvement than rank-25. This indicates that at higher frequencies, maintaining a sufficient rank is critical for capturing the detailed characteristics of the wavefield. 

These results reveal several important insights about the impact of rank size on Meta-LRPINN performance:
\begin{itemize}
    \item \textbf{Higher ranks are essential for high-frequency wavefields}: The results confirm that for high-frequency wavefields (e.g., 12 Hz), smaller ranks fail to effectively capture the detailed characteristics of the wavefield, leading to lower accuracy and slower convergence. This highlights the importance of maintaining sufficient rank for high-frequency applications.
    \item \textbf{Over-parameterization risks with a very high rank}: While increasing the rank generally improves accuracy and convergence, excessively large ranks (e.g., rank-200) introduce redundant parameters, leading to over-parameterization. This can make optimization more challenging, resulting in slightly lower accuracy compared to a more optimal rank, as observed in the comparison between rank-100 and rank-200.
    \item \textbf{Advantages of rank-adaptive reduction}: When comparing these results to the adaptive rank reduction experiments (see Figure \ref{fig12}), we can observe that a rank-100 configuration during meta-training, combined with adaptive rank reduction during meta-testing (e.g., 50\% reduction), outperforms models meta-trained and tested with a fixed rank of 50. This demonstrates the potential of adopting a flexible rank strategy: using an appropriate rank during meta-training and applying rank-adaptive reduction during meta-testing can not only enhance computational efficiency but also maintain competitive accuracy and convergence speed.
\end{itemize}

\subsection{The effect of pruning vs. retaining FEH during meta-testing phase}
In the previous numerical examples section, when we perform the meta-testing phase, we choose to prune the FEH to further simplify the network and reduce computational overhead. To investigate the impact of this decision, we here compare the performance of retaining FEH and pruning FEH during the meta-testing phase. We also use the layered velocity model as a example and evaluate three frequencies (3 Hz, 6 Hz, and 12 Hz) by training separate networks for each configuration. 

\begin{figure}[htbp]
\centering
\includegraphics[width=0.95\textwidth]{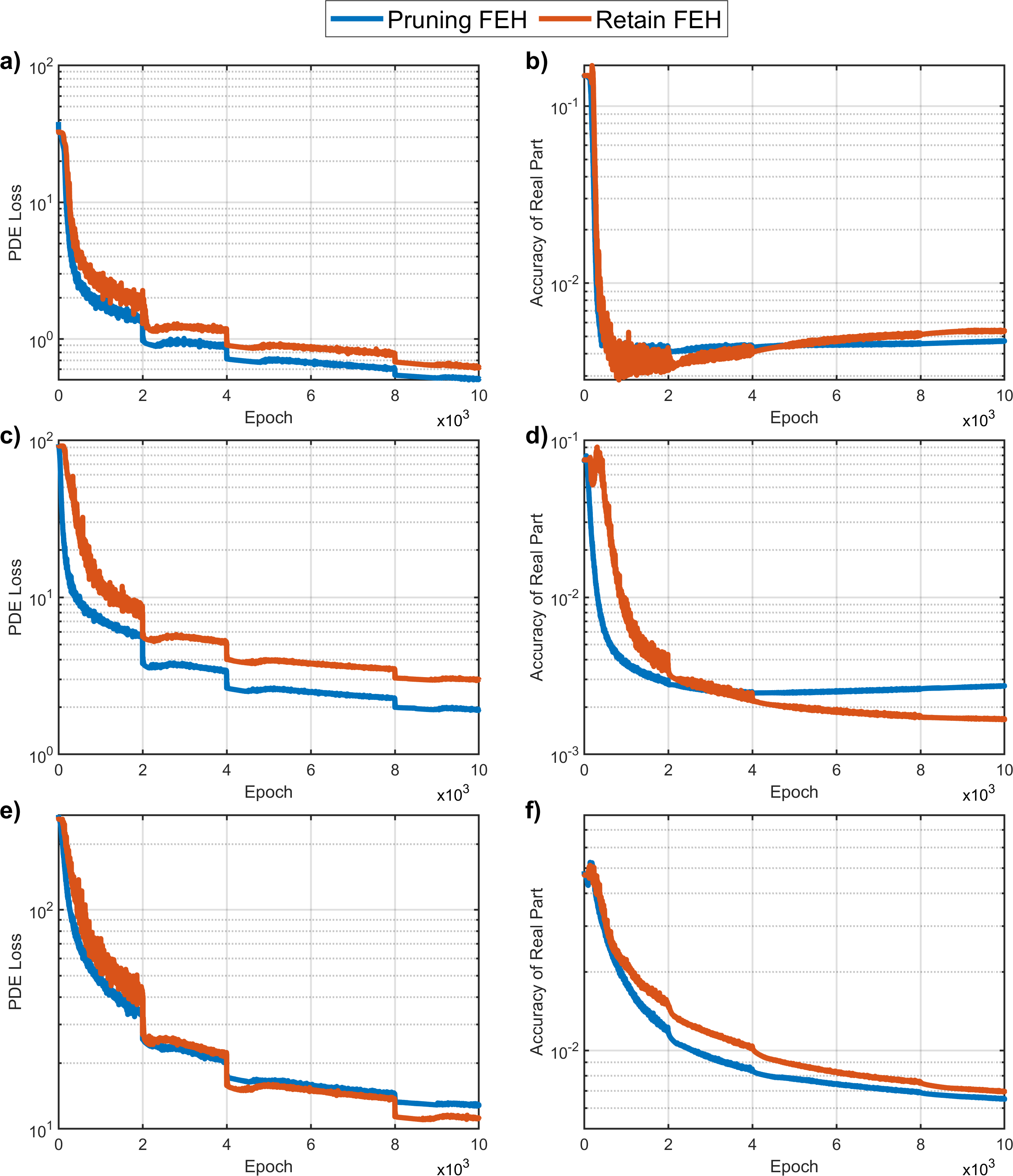}
\caption{Comparison of physical loss and accuracy curves between pruning FEH (blue) and retaining FEH (orange) during the meta-testing phase for 3 Hz (top row), 6 Hz (middle row), and 12 Hz (bottom row). The left column displays physical loss curves, while the right column shows accuracy curves.}
\label{fig18}
\end{figure}

Figure~\ref{fig18} illustrates the physical loss and accuracy curves for both strategies, where the rows from top to bottom correspond to frequencies 3, 6, and 12 Hz, and the left and right columns of each row represent the physical loss and accuracy curves, respectively. The physical loss curves suggest that pruning FEH generally achieves faster convergence compared to retaining FEH in the low- and mid-frequency cases (3 Hz and 6 Hz). Specifically, at both 3 and 6 Hz, the pruned FEH model converges much fast than the retained FEH model. At 12 Hz, the two approaches show similar convergence speeds. For the accuracy curves, at 3 Hz, the accuracy curves for the two strategies nearly overlap. At 6 Hz, pruning FEH achieves faster accuracy improvement in the early training stages, but the retained FEH model eventually achieves slightly higher accuracy in the later stages. At 12 Hz, pruning FEH outperforms the retained FEH model in terms of both convergence speed and final accuracy. 

Therefore, we can see that pruning FEH generally accelerates convergence, particularly in the early stages of training, making it a computationally efficient choice for meta-testing phase. While retaining FEH maintains the network’s full parameterization, the added complexity does not consistently result in better performance. Pruning FEH, on the other hand, reduces computational costs without significant sacrifices and, in some cases, even outperforms retaining FEH. As a result, we recommend directly pruning FEH during the meta-testing phase, as we did in our numerical examples.

\subsection{The performance for out of distribution}
To evaluate the generalization capability of the meta-trained initialization provided by Meta-LRPINN, we test its performance on a frequency outside the range used during the meta-training phase. While the meta-training stage is conducted on frequencies within the range of 2–15 Hz, we select 18 Hz as an out-of-distribution frequency to examine the model's ability to adapt to unseen scenarios. Using the same layered velocity model as in previous experiments, we initialize the meta-testing phase with the meta-trained weights and assess the wavefield representation at 18 Hz. For comparison, we include two benchmarks: Meta-PINN and vanilla PINN. 

\begin{figure}[htbp]
\centering
\includegraphics[width=0.95\textwidth]{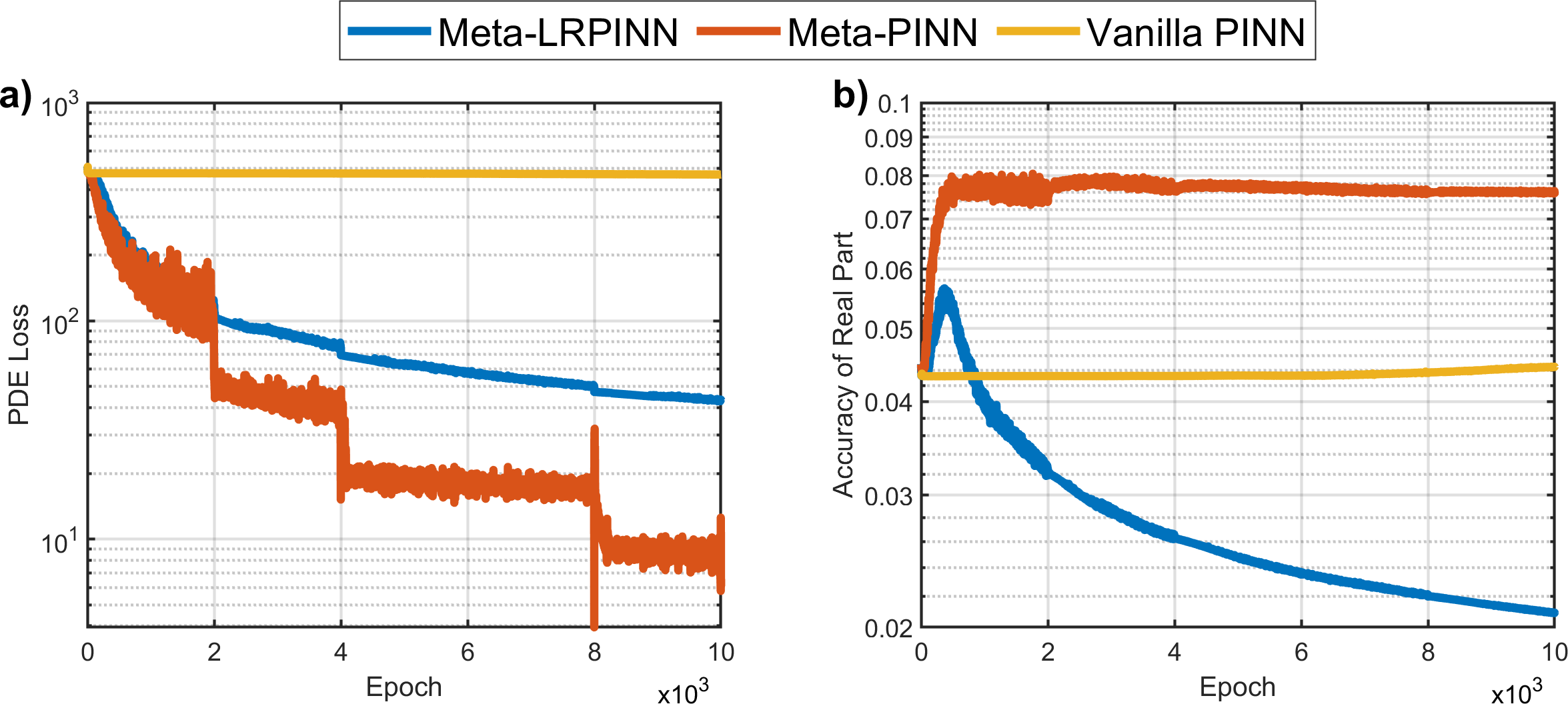}
\caption{Test the performance of our Meta-LRPINN (blue) on an out-of-distribution frequency (18 Hz), where we compare it with Meta-PINN (orange) and vanilla PINN (yellow). (a) The physical loss curves. (b) The accuracy curves.}
\label{fig19}
\end{figure}

Figure~\ref{fig19} compares the performance of our Meta-LRPINN, Meta-PINN, and vanilla PINN, with panel (a) showing the physical loss curves and panel (b) displaying the accuracy curves. We can see that our Meta-LRPINN shows a clear advantage in reducing PDE loss right from the start, indicating that the meta-trained initialization enables the model to quickly establish an effective optimization direction even for a frequency not encountered during meta-training. In terms of accuracy, Meta-LRPINN initially improves rapidly within just a few epochs, but its accuracy subsequently declines. Meta-PINN, despite demonstrating a faster initial decline in PDE loss compared to Meta-LRPINN, ultimately fails to convert that into meaningful accuracy gains. Instead, its accuracy increases over time, implying that the model converges to a trivial solution rather than genuinely improving the wavefield representation. Meanwhile, vanilla PINN’s curves for both PDE loss and accuracy remain essentially flat, underscoring its difficulty in effectively modeling high-frequency wavefields without a tailored initialization and extended optimization effort. Overall, these observations highlight the capability of our Meta-LRPINN in improving adaptability and optimization efficiency at out-of-distribution frequencies.

\section{\textbf{Conclusions}}\label{conclusions}
We proposed Meta-LRPINN, a meta-learning-enhanced low-rank physics-informed neural network (PINN) framework, to address the challenges of modeling multi-frequency wavefields in variable velocity models. We first introduced a singular value decomposition (SVD)-based approach to decompose the hidden layer weights of PINN, significantly reducing the number of parameters, by reducing the rank (the number of singular values). To enhance the model's adaptability to varying frequencies, we proposed an innovative frequency embedding hypernetwork (FEH) that linked the input frequency to the singular values in the SVD decomposition. To improve training efficiency and optimization stability, we used meta-learning to provide a robust initialization. Furthermore, we introduced adaptive rank reduction and FEH pruning during the meta-testing phase to further enhance computational efficiency. Numerical examples demonstrated that Meta-LRPINN achieved faster convergence and higher accuracy than Meta-PINN, which did not include the SVD decomposition or the FEH, as well as vanilla PINN, with superior adaptability to varying frequencies and velocity models. The results also highlighted that adaptive rank reduction maintained competitive accuracy while significantly reducing the computational cost, and pruning FEH accelerated convergence without compromising performance. These findings demonstrate Meta-LRPINN as a robust and scalable framework for seismic wavefield representation.

\section*{\textbf{Acknowledgments}}
This publication is based on work supported by the King Abdullah University of Science and Technology (KAUST). The authors thank the DeepWave sponsors fort heir support. This work utilized the resources of the Supercomputing Laboratory at King Abdullah University of Science and Technology (KAUST) in Thuwal, Saudi Arabia.
\section*{\textbf{Code and Data Availability}}
The data and accompanying codes that support the findings of this study are available at: 
\url{https://github.com/DeepWave-KAUST/Meta-LRPINN}. (During the review process, the repository is private. Once the manuscript is accepted, we will make it public.)

\bibliographystyle{unsrtnat}
\bibliography{references}

\end{document}